\documentclass[letterpaper]{article} 
\usepackage{aaai2027}  
\usepackage[hyphens]{url}  
\usepackage{graphicx} 
\graphicspath{{figures/}} 
\usepackage{natbib}  
\usepackage{caption} 
\usepackage{algorithm}
\usepackage{algorithmic}
\usepackage{tabularx}
\usepackage{array}
\usepackage{amsmath}
\usepackage{amssymb}

\usepackage{newfloat}
\usepackage{listings}
\DeclareCaptionStyle{ruled}{labelfont=normalfont,labelsep=colon,strut=off} 
\floatstyle{ruled}
\newfloat{listing}{tb}{lst}{}
\floatname{listing}{Listing}

\usepackage{booktabs}
\renewcommand{\topfraction}{0.92}
\renewcommand{\bottomfraction}{0.75}
\renewcommand{\textfraction}{0.07}
\renewcommand{\floatpagefraction}{0.75}

\title{CoMedBench: A Multi-Source Benchmark of Synthetic Medical Data Fidelity and Downstream Utility}

\author{
    Akanta Das\textsuperscript{\rm 1},
    Farhad Al-Amin Dipto\textsuperscript{\rm 1},
    Mrinmoy Sarkar Anto\textsuperscript{\rm 1},\\
    David Rehkopf\textsuperscript{\rm 2},
    Ayin Vala\textsuperscript{\rm 2},
    Tanmoy Sarkar Pias\textsuperscript{\rm 2}
}

\affiliations{
    \textsuperscript{\rm 1}Bangladesh University of Engineering and Technology, Dhaka, Bangladesh\\
    \textsuperscript{\rm 2}School of Medicine, Stanford University, Stanford, California, United States
}
\begin{document}

\maketitle

\begin{abstract}
Access to clinical data is essential for developing reliable healthcare machine learning systems, but direct use of electronic health records is constrained by privacy regulation, institutional review, data-use agreements, and the risk of re-identification. Synthetic data promises a practical alternative: it can preserve useful statistical and clinical structure while reducing exposure of sensitive patient records. Yet the evidence for synthetic data utility in healthcare remains fragmented. Prior studies often evaluate a single generator, one dataset, or a narrow downstream task, making it difficult to know when synthetic data can support model development and when it fails to preserve task-critical signal. We introduce \textbf{CoMedBench}, a reproducible benchmark that evaluates a family of generators under a common clinical-validity framework and one shared training and evaluation engine, spanning static tabular and temporal downstream tasks on established critical-care datasets. In total the benchmark spans 37 dataset-task pairs across two modalities consists of 20 static tabular and 17 temporal ICU time-series-drawn from seven public data sources: three intensive-care databases (\textbf{MIMIC-III, MIMIC-IV, and eICU}) together with the\textbf{ UCI }Machine Learning Repository, the CDC \textbf{BRFSS} diabetes cohort (2015), \textbf{NHANES} (1999-2014), and the pycox survival datasets ( \textbf{GBSG} and  \textbf{METABRIC}). The benchmark evaluates both statistical fidelity and task utility by comparing models trained and tested across real and synthetic data. In these settings, synthetic training data preserves most of the downstream signal: on tabular tasks the reference generator CoMed-CTGAN retains a mean AUROC utility (the synthetic-to-real performance ratio) of \textbf{$90.6\%$}, rising to \textbf{$97.3\%$} for the strongest generator, CoMed-TVAE. Temporal ICU tasks are harder and more generator-sensitive: CoMed-CTGAN retains \textbf{$81.6\%$} (AUROC) and only \textbf{$64.0\%$} under the imbalance-sensitive AUPRC, whereas CoMed-TVAE still retains \textbf{${\sim}95\%$} (AUROC). By organizing evaluation around downstream clinical prediction rather than visual similarity alone, the benchmark clarifies where synthetic data is useful, where temporal structure remains a bottleneck, and how future generators should be assessed for healthcare AI.
\end{abstract}

\section{Introduction}

\begin{figure}[t]
\centering
\includegraphics[width=\columnwidth]{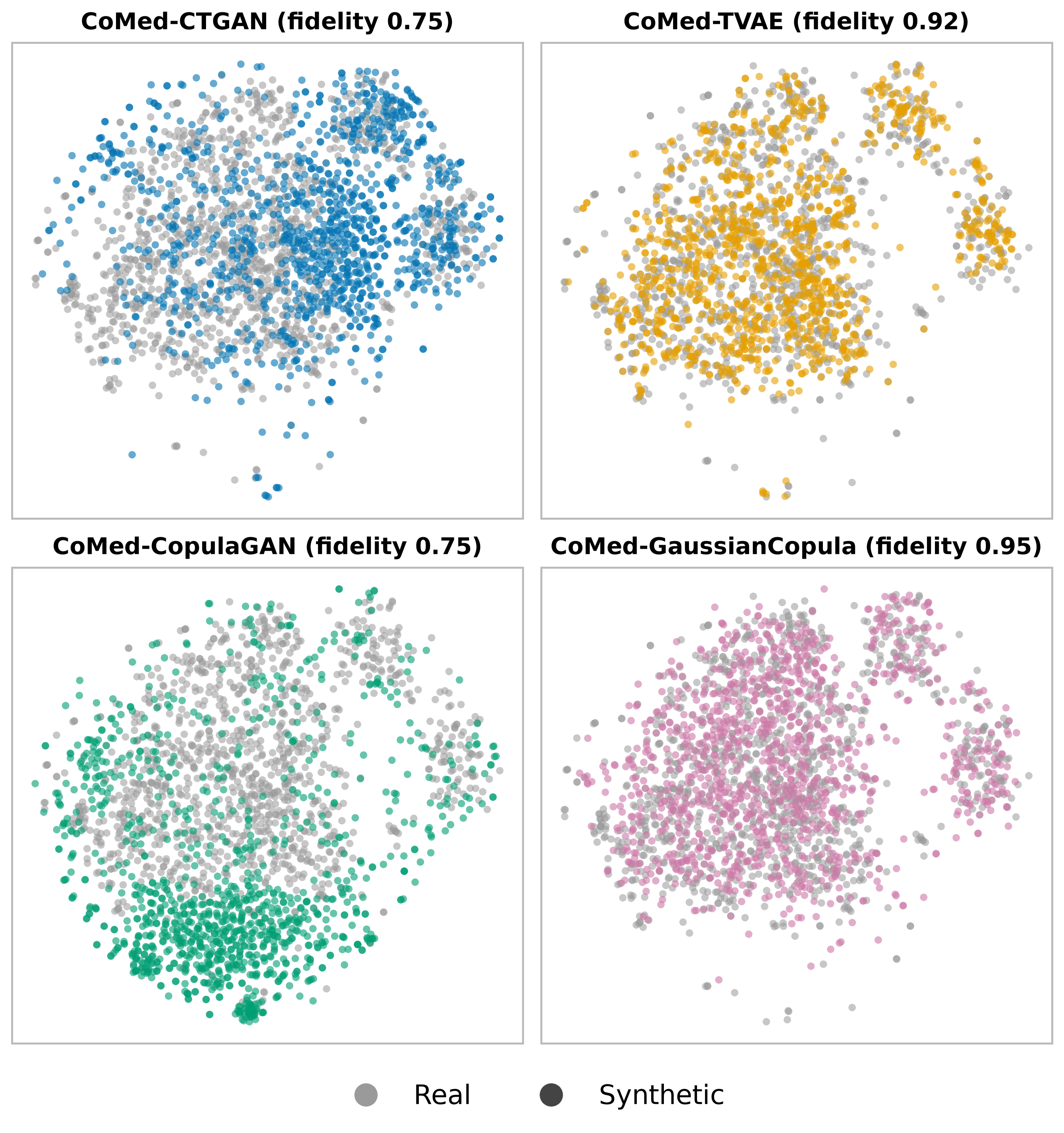}
\caption{Two-dimensional t-SNE (perplexity $30$, standardized features) of real
vs.\ synthetic ICU records for the MIMIC-III heart-failure cohort. Generators differ
markedly in how well synthetic data covers the real manifold: higher-fidelity models
(CoMed-TVAE, CoMed-GC) mix throughout the real points, while lower-fidelity models
(CoMed-CTGAN, CoMed-CopulaGAN) leave real regions under-covered. This gap between
visual/statistical similarity and task usefulness motivates our benchmark.}
\label{fig:tsne}
\end{figure}

Modern healthcare AI depends on large, representative, and carefully curated clinical datasets \citep{habehh2021machine}. Electronic health records (EHRs), intensive-care databases, laboratory measurements, and longitudinal monitoring streams contain signals that support mortality prediction, readmission modeling, length-of-stay estimation, and phenotyping. However, the same data are hard to share: clinical datasets include protected health information, rare diagnoses, and combinations of demographic and clinical attributes that can make patients identifiable even after conventional de-identification. As a result, many healthcare machine learning studies depend on a small number of public resources such as MIMIC-III, MIMIC-IV, and eICU, while large institutional datasets remain inaccessible to most researchers \citep{johnson2016mimic,johnson2023mimiciv,pollard2018eicu}.

\begin{figure*}[t]
\centering
\includegraphics[width=\textwidth]{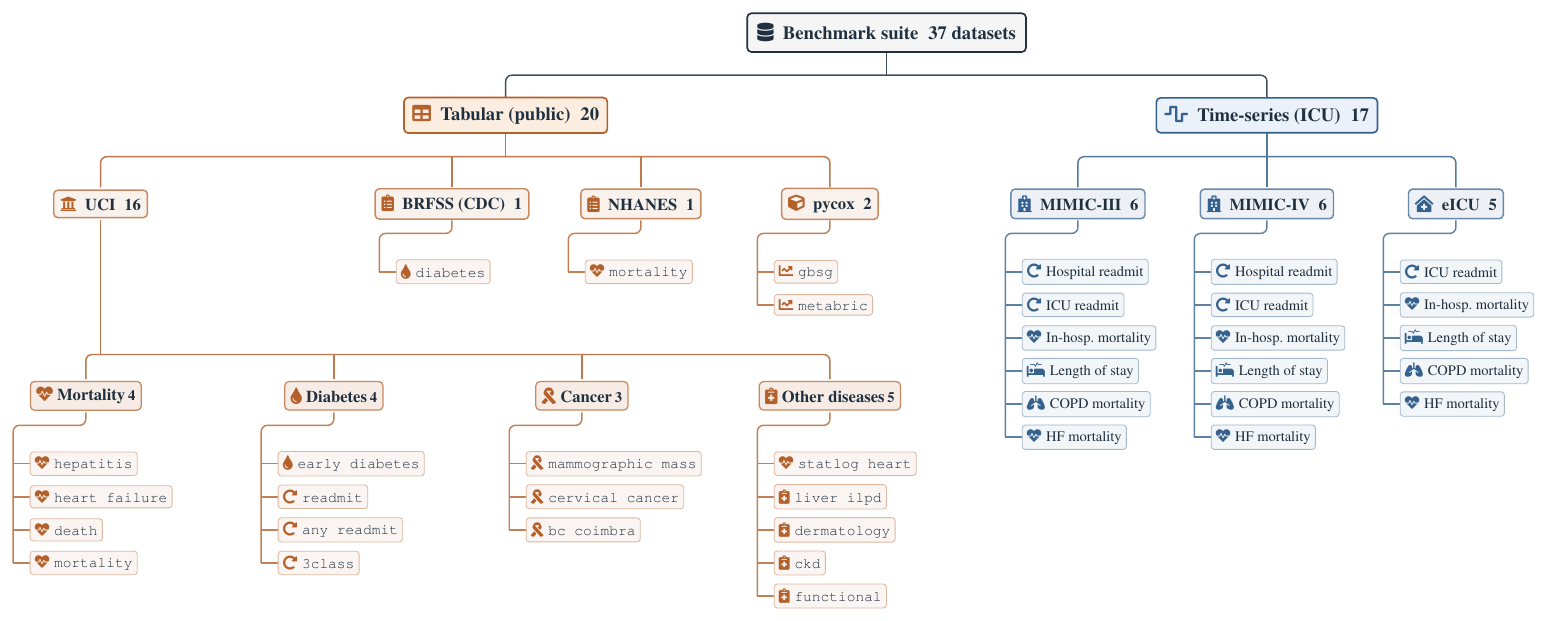}
\caption{Benchmark coverage. The 37 dataset-task pairs split by modality (tabular/public vs.\
time-series/ICU) and by source; the public UCI datasets are further grouped into clinical
subclasses (mortality, diabetes, cancer, and other diseases), while each ICU database
(MIMIC-III, MIMIC-IV, eICU) contributes mortality, readmission, and length-of-stay tasks.}
\label{fig:coverage}
\end{figure*}

Synthetic clinical data has emerged as a promising way to reduce this access barrier. Instead of releasing original patient records, a data holder trains a generative model and releases artificial records intended to approximate the distribution of the source data. If successful, synthetic data can enable algorithm prototyping, external benchmarking, software testing, and multi-institutional collaboration without exposing raw records.

The central challenge is that synthetic data must be useful, not merely realistic. A generator can match marginal distributions while distorting rare outcomes, comorbidity structure, medication-laboratory dependencies, or temporal dynamics. These distortions are harmful for downstream tasks because the predictive signal often lies in minority classes, rare events, and time-dependent trajectories \citep{Pias2025}. A synthetic dataset that appears faithful under aggregate statistics may therefore fail when used to train a model for a clinically relevant task; Figure~\ref{fig:tsne} illustrates this gap.

Existing synthetic-data evaluations in healthcare report fidelity, privacy, and utility, but utility is often measured under different datasets, models, metrics, and task definitions, which makes comparison difficult. Researchers thus still lack a practical answer to a basic question: for which healthcare downstream tasks can synthetic data act as a useful substitute for real clinical data?

This paper addresses that gap with CoMedBench, a benchmark built to make synthetic-data evaluation controlled and comparable. CoMedBench applies a model-agnostic clinical-validity layer and wraps a family of generators in a single training pipeline so that the generator is the only component that varies between runs, and evaluates each synthetic table through two complementary scenarios: statistical fidelity, which measures how closely generated records resemble the source distribution, and downstream utility, which measures how well models trained on synthetic data transfer to real test data. This design separates apparent realism from practical usefulness.

Our work makes the following contributions:
\begin{itemize}
    \item \textbf{Controlled evaluation under domain validity.} We evaluate four generator families under a common clinical-validity framework and one shared training and evaluation pipeline, so performance is compared without confounding differences in preprocessing, structural validation, or downstream evaluation.
    \item \textbf{A multi-source clinical benchmark.} 37 dataset-task pairs-20 static tabular and 17 temporal ICU time-series from seven public sources, each evaluated on statistical fidelity and downstream utility.
    \item \textbf{Synthetic data preserves downstream signal, but unevenly.} The strongest generator, CoMed-TVAE, retains ${\sim}97\%$ of real-data AUROC on tabular tasks and ${\sim}95\%$ on time-series; utility is high on static tabular tasks but degrades on temporal, imbalanced ICU tasks (AUPRC utility as low as ${\sim}64\%$), and no single generator dominates across the evaluation settings.
    \item \textbf{Fidelity is a necessary but insufficient proxy.} Overall statistical fidelity correlates with downstream utility only moderately (Overall $r=0.67$, $\rho=0.74$); utility must therefore be measured directly, with rare-event temporal ICU tasks the main bottleneck.
\end{itemize}

\section{Related Works}

\begin{figure*}[t]
\centering
\includegraphics[width=\textwidth]{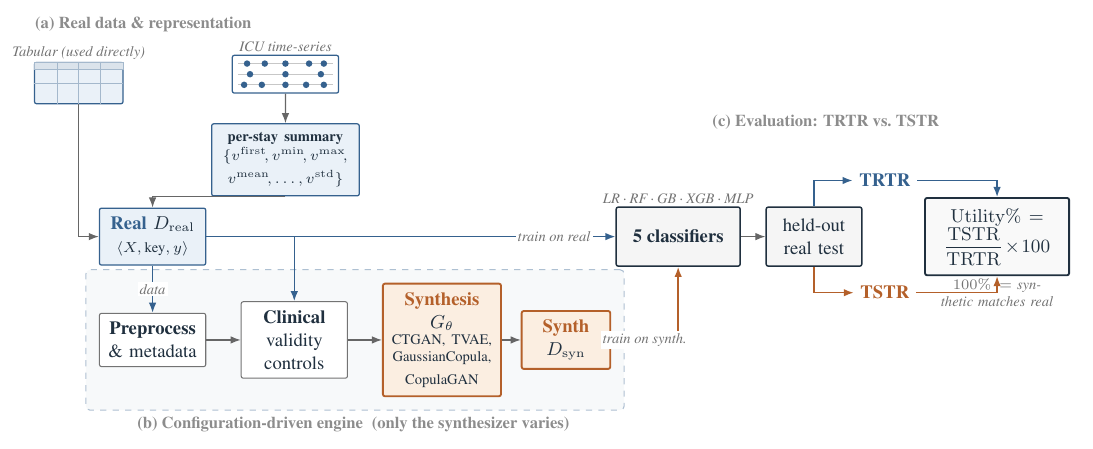}
\caption{The CoMedBench pipeline. Real data (used directly as a flat table, or an
ICU stay's multivariate series summarized into a per-stay feature vector) drives a single
configuration-driven engine-preprocessing and metadata, clinical-validity controls, and a
synthesizer $G_\theta$-whose only stage that changes across runs is the
synthesizer. The resulting synthetic table $D_{\mathrm{syn}}$ and the real table
$D_{\mathrm{real}}$ are then compared by training the same five classifiers on each and
testing on an \emph{identical} held-out real set (TRTR vs.\ TSTR), yielding the utility ratio.}
\label{fig:pipeline}
\end{figure*}

\subsection{Synthetic Data in Tabular and Time-Series Data}

Synthetic data for structured data can be grouped into single-table, multi-table, and time-series synthesis. Single-table methods generate one flat table with mixed categorical and continuous columns, focusing on distributional fidelity, privacy, and machine-learning utility; examples include CTGAN-style models, differentially private normalizing flows, diffusion models, and autoencoder-based tabular generators \citep{xu2019ctgan,patki2016sdv,lee2022dpflows}. These methods suit static clinical cohorts but may lose key relationships when a relational EHR database is flattened into one table.

Multi-table synthesis instead preserves dependencies across interconnected tables (e.g., patient-admission, diagnosis-procedure, and laboratory-medication relationships), better reflecting real-world healthcare databases in which downstream labels depend on information distributed across multiple tables. Such relational methods are important for reconstructing realistic EHR structure rather than only generating independent rows.

Time-series synthesis focuses on sequential records where order, missingness, irregular sampling, and temporal dynamics are clinically meaningful. This direction is especially relevant for ICU tasks because prior work using models such as GRU-D and established MIMIC-III benchmarks has shown that temporal trends and missingness patterns are predictive for mortality, decompensation, length of stay, and phenotyping \citep{che2018grud,harutyunyan2019multitask}.

\subsection{Synthetic Data in Downstream Tasks}

Synthetic data is most useful when it supports downstream tasks, not only when it looks statistically similar to real data. A common evaluation is train-on-synthetic, test-on-real (TSTR): models trained on generated data are evaluated on a held-out real test set, directly measuring whether synthetic data preserves transferable predictive signal.

Established MIMIC-III and MIMIC-IV benchmarks provide common structured and time-series tasks-mortality, readmission, length-of-stay, and phenotyping-for evaluating whether synthetic clinical data preserves useful predictive signal \citep{harutyunyan2019multitask,che2018grud,gupta2022mimicivpipeline,gupta2022extensive,bui2024benchmarking,li2021prediction,peng2022interpretable,qiu2022interpretable,lin2019analysis,adisa2026integratedframeworkexplainablefair,kakadiaris2023evaluatingfairnessmimicivdataset}. These tasks motivate our benchmark because synthetic data may perform well on static tabular prediction while failing to preserve temporal patterns or rare-event signals.

\subsection{Synthetic Data in Healthcare AI}

Public datasets such as MIMIC-III, MIMIC-IV, and eICU have enabled critical-care research, but broad access to institutional EHR data remains limited \citep{johnson2016mimic,johnson2023mimiciv,pollard2018eicu}.

Synthetic healthcare data can support prototyping, testing, education, and benchmarking without exposing raw patient records \citep{Chen2021,Tucker2020}. However, privacy protection alone is insufficient: generated data must also preserve clinical structure such as outcome prevalence, comorbidity patterns, medication-laboratory relationships, and temporal deterioration. This motivates our benchmark, which evaluates whether synthetic data supports established downstream tasks rather than only matching broad statistical properties.

\section{Methodology}

%

\subsection{Overview of the CoMedBench Framework}
 
CoMedBench treats synthetic clinical-data generation and evaluation as a single,
configuration-driven procedure applied uniformly to every (dataset~$\times$~task)
pair. Each experiment is defined by a compact task specification: the input
feature matrix, the primary-key column, the prediction label and the feature-typing
rule. One shared engine consumes this
specification and executes four stages
(Figure~\ref{fig:pipeline}):
(1)~\textit{preprocessing and metadata}, which loads the flat feature matrix,
imputes missing values, and assigns each column an explicit type;
(2)~\textit{clinical-validity controls}, a model-agnostic layer that enforces
schema and task-specific validity rules consistently across every generator;
(3)~\textit{synthesis}, which fits a generator $G_\theta$ and
samples a synthetic table $D_{\mathrm{syn}}$ whose label distribution matches the
real class proportions; and
(4)~\textit{evaluation}, which scores $D_{\mathrm{syn}}$ against the real data
$D_{\mathrm{real}}$ along two axes, statistical \emph{fidelity} and
downstream machine-learning \emph{utility} (TRTR vs.\ TSTR). Because the same four
stages are applied to every dataset, task, and generator, only the synthesis stage (or a
single ablated setting) changes between runs, making the study a controlled
benchmark rather than a comparison of loosely coupled implementations.
 
\subsection{Datasets, Tasks, and Representation}
 
We evaluate on three intensive-care databases - MIMIC-III, MIMIC-IV, and
eICU \citep{johnson2016mimic,johnson2023mimiciv,pollard2018eicu}, together with a collection of public clinical and tabular datasets
\citep{strack2014impact,patricio2018using,ramana2011critical,chicco2020machine,katzman2018deepsurv,xie2019building,11186191}. Tasks are binary or
multi-class prediction: in-hospital mortality, ICU and hospital 30-day
readmission, ICU length of stay, and disease/outcome classification.
Figure~\ref{fig:coverage} shows how the suite breaks down by modality, source, and task.
 
Every task is modeled as a single flat table (one row per patient or stay). In the
\emph{tabular} representation the dataset is already a flat feature matrix and is
used directly. In the \emph{time-series-to-tabular} representation, each ICU stay's
multivariate time series is summarized into a per-stay feature vector: each
clinical variable $v$ is represented by $v_{\mathrm{first}}$, $v_{\min}$,
$v_{\max}$, and $v_{\mathrm{mean}}$ (and, where available, $v_{\mathrm{last}}$,
$v_{\mathrm{median}}$, $v_{\mathrm{std}}$) alongside demographic and contextual
features, yielding one fixed-width row per stay.
 
\subsection{Preprocessing and Metadata}
 
Each table is loaded and assigned a primary key, a native identifier where one
exists (e.g., stay id) or a surrogate key otherwise; rows with a missing
label are dropped. We build single-table metadata assigning every column an explicit
type: the key as an identifier (regenerated at sampling time), the label as
categorical, and each feature as categorical or numerical.
Missing values follow a domain-aware imputation policy, and low-signal columns are
pruned \citep{tang2020democratizing}. High-cardinality categorical fields are encoded so that generators emit only
categories observed in training. 


\subsection{Clinical Validity Controls}

To ensure that generated records satisfy basic clinical and structural validity
requirements, we apply a model-agnostic domain-validity layer consistently across
all evaluated generators. The layer uses schema information and task-specific
validity rules to prevent clinically impossible or structurally invalid outputs.
This study evaluates the effect of applying the same validity layer across
generator families.

\subsection{Synthesis: Generator Families}

The same clinical-validity layer is applied identically to four single-table
generators from distinct families, so the synthesis stage is the only component
that changes between runs. \textbf{CoMed-CTGAN} is a conditional tabular generative adversarial network with
mode-specific normalization and training-by-sampling for imbalanced discrete columns
\citep{xu2019ctgan}. \textbf{CoMed-TVAE}
is a tabular variational autoencoder, \textbf{CoMed-CopulaGAN} a copula-augmented
conditional GAN, and \textbf{CoMed-GaussianCopula} a statistical Gaussian-copula model with no
training epochs. The GAN-based and TVAE-based generators use standard published
configurations.
Because
the labels are imbalanced, generation is
\emph{label-conditioned}: the synthetic label distribution matches the real class
proportions, and the number of synthetic rows equals the real dataset size.
 
\subsection{Evaluation}
 
We evaluate every synthetic table against the real table on two axes: statistical
fidelity and downstream machine-learning utility (TRTR/TSTR). For brevity write
$R=D_{\mathrm{real}}$ and $S=D_{\mathrm{syn}}$, and let $\mathcal{C}$ be the set of
modeled columns, partitioned into numerical columns $\mathcal{C}_{\mathrm{num}}$ and
categorical columns $\mathcal{C}_{\mathrm{cat}}$, with $\mathcal{P}$ the set of
column pairs.
 
\begin{figure}[t]
\centering
\includegraphics[width=\columnwidth]{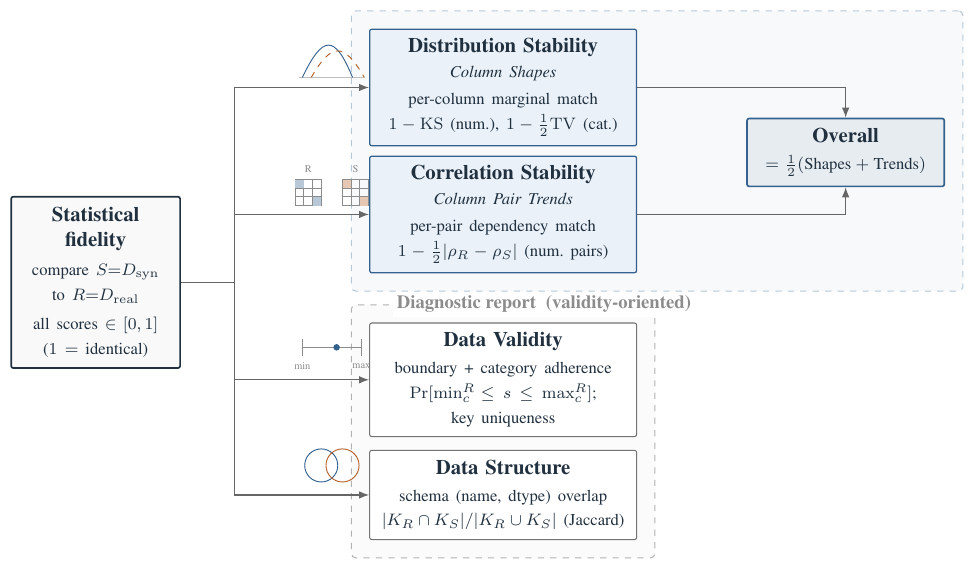}
\caption{The CoMedBench fidelity suite. A \emph{Quality} score combines Distribution
Stability (per-column marginals) and Correlation Stability (column-pair trends); a
\emph{Diagnostic} score adds Data Validity and Data Structure. All scores lie
in $[0,1]$ ($1=$ identical).}
\label{fig:metrics}
\end{figure}
 
\paragraph{Statistical fidelity.}
Figure~\ref{fig:metrics} organizes the fidelity metrics used below; each is computed
by CoMedBench from standard column and column-pair statistics.
\emph{Distribution Stability} averages a per-column marginal score: the
Kolmogorov-Smirnov complement for numerical columns and the total-variation
complement for categorical columns. For a numerical column $c$ with empirical CDFs
$\hat{F}^{R}_{c},\hat{F}^{S}_{c}$,
\begin{equation}
\mathrm{KSC}(c)=1-\sup_{x}\bigl|\hat{F}^{R}_{c}(x)-\hat{F}^{S}_{c}(x)\bigr|,
\end{equation}
and for a categorical column $c$ with category set $\mathcal{A}_c$ and relative
frequencies $R_c(a),S_c(a)$,
\begin{equation}
\mathrm{TVC}(c)=1-\tfrac{1}{2}\sum_{a\in\mathcal{A}_c}\bigl|R_c(a)-S_c(a)\bigr|.
\end{equation}
Distribution Stability is their mean over columns,
\begin{equation}
\mathrm{Shapes}=\frac{1}{|\mathcal{C}|}\!\left(\sum_{c\in\mathcal{C}_{\mathrm{num}}}\!\mathrm{KSC}(c)+\sum_{c\in\mathcal{C}_{\mathrm{cat}}}\!\mathrm{TVC}(c)\right).
\end{equation}
\emph{Correlation Stability} averages a per-pair
dependency score. For a numerical pair $(a,b)$ with Pearson correlations
$\rho^{R}_{ab},\rho^{S}_{ab}$,
\begin{equation}
\mathrm{CS}(a,b)=1-\frac{\bigl|\rho^{R}_{ab}-\rho^{S}_{ab}\bigr|}{2},
\end{equation}
and for any pair involving a categorical column (numerical columns discretized into
bins) with joint relative frequencies $R_{ab}(i,j),S_{ab}(i,j)$,
\begin{equation}
\mathrm{CT}(a,b)=1-\tfrac{1}{2}\sum_{i,j}\bigl|R_{ab}(i,j)-S_{ab}(i,j)\bigr|.
\end{equation}
Correlation Stability is
\begin{equation}
\mathrm{Trends}=\frac{1}{|\mathcal{P}|}\sum_{(a,b)\in\mathcal{P}}m(a,b),
\end{equation}
where
\begin{equation}
m(a,b)=\begin{cases}\mathrm{CS}(a,b), & a,b\in\mathcal{C}_{\mathrm{num}},\\[2pt]\mathrm{CT}(a,b), & \text{otherwise,}\end{cases}
\end{equation}
and the \emph{Overall} quality score is the mean of the two properties,
\begin{equation}
\mathrm{Overall}=\tfrac{1}{2}\bigl(\mathrm{Shapes}+\mathrm{Trends}\bigr).
\end{equation}
The diagnostic report additionally measures \emph{Data Validity} and
\emph{Data Structure}. Data Validity averages a per-column adherence score:
boundary adherence for numerical columns and category adherence for categorical
columns,
\begin{align}
\mathrm{BA}(c)&=\frac{1}{|S|}\sum_{s\in S_c}\mathbf{1}\!\left[\min_c R\le s\le\max_c R\right],\\
\mathrm{CA}(c)&=\frac{1}{|S|}\sum_{s\in S_c}\mathbf{1}\!\left[s\in\mathcal{A}^{R}_{c}\right],
\end{align}
with primary keys scored by uniqueness
$\mathrm{KU}=1-\tfrac{1}{|S|}\#\{\text{missing or duplicate synthetic keys}\}$.
Data Structure is the Jaccard overlap of the (column name, dtype) schemas
$\mathcal{K}^{R},\mathcal{K}^{S}$,
\begin{equation}
\mathrm{Structure}=\frac{|\mathcal{K}^{R}\cap\mathcal{K}^{S}|}{|\mathcal{K}^{R}\cup\mathcal{K}^{S}|}.
\end{equation}
All fidelity scores lie in $[0,1]$ ($1$ = identical).
 
\paragraph{Machine-learning utility (TRTR/TSTR).} For each task we train five
classifiers- Logistic Regression, Random Forest, Gradient Boosting, XGBoost, and an
MLP, under two regimes: \emph{TRTR} (Train Real, Test Real; upper-bound reference)
and \emph{TSTR} (Train Synthetic, Test Real, on the \emph{same} held-out real test
set). 
We summarize the gap as a Utility ratio,
\begin{equation}
\mathrm{Utility\%}=\frac{\mathrm{AUC}_{\mathrm{TSTR}}}{\mathrm{AUC}_{\mathrm{TRTR}}}\times 100,
\end{equation}
so $100\%$ means synthetic-trained models match real-trained ones.

\section{Results}
 
We report statistical fidelity across the four generators, downstream utility, and
the relationship between them. Fidelity is measured for all four synthesizers, with a per-dataset breakdown in
Table~\ref{tab:fidelity_heatmaps}. Downstream utility is reported per dataset for all four synthesizers in
Table~\ref{tab:util} (AUROC and AUPRC),
each aggregated over the five classifiers.

\begin{table*}[t]
\centering
\small
\setlength{\tabcolsep}{6pt}
\resizebox{\textwidth}{!}{%
\begin{tabular}{llcccc}
\toprule
Dataset & Task & CoMed-CTGAN & CoMed-TVAE & CoMed-CopulaGAN & CoMed-GaussianCopula \\
\midrule
UCI Cervical Cancer & Cervical Cancer        & 0.867 ($+0.098$) & 0.913 ($-0.023$) & 0.828 ($+0.044$) & 0.877 ($+0.100$) \\
MIMIC-IV  & Heart Failure Mortality          & 0.825 ($+0.038$) & 0.883 ($-0.007$) & 0.822 ($+0.030$) & 0.866 ($-0.016$) \\
MIMIC-III & Heart Failure Mortality          & 0.754 ($+0.016$) & 0.915 ($-0.006$) & 0.749 ($+0.013$) & 0.955 ($+0.000$) \\
MIMIC-III & Length-of-Stay                   & 0.840 ($+0.006$) & 0.895 ($-0.013$) & 0.841 ($+0.010$) & 0.871 ($-0.040$) \\
MIMIC-IV  & Length-of-Stay                   & 0.916 ($+0.004$) & 0.902 ($-0.001$) & 0.913 ($+0.003$) & 0.926 ($+0.000$) \\
\bottomrule
\end{tabular}%
}
\caption{Effect of the clinical-validity layer on statistical fidelity. Each cell reports Overall fidelity with the validity layer, and in
parentheses the change relative to the same generator without it
(positive $=$ the validity layer improves fidelity). The layer improves fidelity for the GAN-based CoMed-CTGAN and CoMed-CopulaGAN on nearly every task (up to $+0.098$), while the CoMed-TVAE and CoMed-GaussianCopula models change only marginally ($|\Delta|\le 0.04$; slightly negative for CoMed-TVAE).}
\label{tab:constraint_fidelity}
\end{table*}
 
\begin{table*}[t]
\centering
\footnotesize

\begin{minipage}[t]{0.49\textwidth}
\vspace{0pt}
\centering
\fontsize{7.2pt}{8.2pt}\selectfont
\setlength{\tabcolsep}{1.4pt}
\renewcommand{\arraystretch}{1.0}
\begin{tabularx}{\linewidth}{
@{}
>{\raggedright\arraybackslash}p{0.28\linewidth}
>{\raggedright\arraybackslash}X
*{4}{>{\centering\arraybackslash}p{0.095\linewidth}}
@{}
}
\toprule
\multicolumn{6}{c}{\textbf{Tabular}} \\
\midrule
\textbf{Dataset} & \textbf{Task} &
\textbf{CoMed\-CTGAN } & \textbf{CoMed\-TVAE } & \textbf{CoMed\-CG } & \textbf{CoMed\-GC } \\

\midrule
Breast Cancer Coimbra & Breast Cancer & 0.744 & 0.893 & 0.760 & \textbf{0.895} \\
CDC BRFSS & Diabetes & \textbf{0.970} & 0.940 & 0.940 & 0.849 \\
GBSG (pycox) & Breast Cancer Survival & 0.935 & \textbf{0.967} & 0.909 & 0.883 \\
METABRIC (pycox) & Breast Cancer Survival & 0.953 & \textbf{0.979} & 0.929 & 0.907 \\
SUPPORT2 & Death & 0.828 & 0.866 & 0.825 & \textbf{0.867} \\
SUPPORT2 & Functional Outcome & 0.816 & \textbf{0.879} & 0.847 & 0.823 \\
SUPPORT2 & Mortality & 0.846 & \textbf{0.873} & 0.810 & 0.863 \\
UCI Cervical Cancer & Cervical Cancer & 0.867 & \textbf{0.913} & 0.828 & 0.877 \\
UCI Kidney Disease & Chronic Kidney Disease & 0.846 & \textbf{0.870} & 0.838 & 0.819 \\
UCI Dermatology & Dermatological Disease & 0.905 & \textbf{0.966} & 0.899 & 0.841 \\
UCI Diabetes & 30-day Readmission & 0.794 & 0.743 & 0.788 & \textbf{0.825} \\
UCI Diabetes & Any-Readmission & 0.795 & 0.765 & 0.786 & \textbf{0.827} \\
UCI Diabetes & Readmission (3-class) & 0.785 & 0.765 & 0.763 & \textbf{0.824} \\
UCI Early Diabetes & Early Diabetes Risk & 0.886 & \textbf{0.958} & 0.863 & 0.922 \\
UCI Heart Failure & Heart Failure Mortality & 0.880 & \textbf{0.929} & 0.835 & 0.922 \\
UCI Hepatitis & Hepatitis Mortality & 0.896 & \textbf{0.940} & 0.884 & 0.830 \\
UCI Indian Liver & Liver Disease & 0.830 & \textbf{0.936} & 0.855 & 0.918 \\
UCI Mammographic & Mass Malignancy & 0.951 & \textbf{0.977} & 0.940 & 0.839 \\
UCI Statlog Heart & Heart Disease & 0.903 & \textbf{0.946} & 0.863 & 0.829 \\
NHANES & Mortality & 0.767 & \textbf{0.944} & 0.804 & 0.944 \\
\bottomrule
\end{tabularx}
\end{minipage}
\hfill
\begin{minipage}[t]{0.49\textwidth}
\vspace{0pt}
\centering
\fontsize{7.2pt}{8.2pt}\selectfont
\setlength{\tabcolsep}{1.4pt}
\renewcommand{\arraystretch}{1.0}
\begin{tabularx}{\linewidth}{
@{}
>{\raggedright\arraybackslash}p{0.28\linewidth}
>{\raggedright\arraybackslash}X
*{4}{>{\centering\arraybackslash}p{0.095\linewidth}}
@{}
}
\toprule
\multicolumn{6}{c}{\textbf{Time-series}} \\
\midrule
\textbf{Dataset} & \textbf{Task} &
\textbf{CoMed\-CTGAN} & \textbf{CoMed\-TVAE} & \textbf{CoMed\-CG} & \textbf{CoMed\-GC} \\

\midrule
MIMIC-III & COPD Mortality & 0.796 & 0.893 & 0.791 & \textbf{0.904} \\
MIMIC-III & Heart Failure Mortality & 0.754 & 0.915 & 0.749 & \textbf{0.955} \\
MIMIC-III & Hospital Readmission & 0.829 & \textbf{0.892} & 0.829 & 0.875 \\
MIMIC-III & Length-of-Stay & 0.840 & \textbf{0.895} & 0.841 & 0.871 \\
MIMIC-III & ICU Readmission & 0.836 & \textbf{0.892} & 0.832 & 0.875 \\
MIMIC-III & Hospital Mortality & 0.846 & \textbf{0.873} & 0.847 & 0.815 \\
MIMIC-IV & COPD Mortality & 0.805 & \textbf{0.885} & 0.810 & 0.858 \\
MIMIC-IV & Heart Failure Mortality & 0.825 & \textbf{0.883} & 0.822 & 0.866 \\
MIMIC-IV & Hospital Readmission & 0.845 & \textbf{0.928} & 0.887 & 0.915 \\
MIMIC-IV & Length-of-Stay & 0.916 & 0.902 & 0.913 & \textbf{0.926} \\
MIMIC-IV & ICU Mortality & 0.856 & \textbf{0.890} & 0.857 & 0.845 \\
MIMIC-IV & ICU Readmission & 0.754 & 0.894 & 0.749 & \textbf{0.912} \\
eICU & COPD Mortality & 0.800 & 0.840 & 0.809 & \textbf{0.865} \\
eICU & Heart Failure Mortality & 0.832 & 0.860 & 0.819 & \textbf{0.872} \\
eICU & Length-of-Stay & 0.857 & 0.863 & \textbf{0.865} & 0.863 \\
eICU & Mortality & 0.907 & 0.884 & \textbf{0.915} & 0.860 \\
eICU & Readmission & 0.863 & \textbf{0.890} & 0.854 & 0.885 \\
\bottomrule
\end{tabularx}
\end{minipage}
\caption{Per-dataset Overall fidelity for the four generators on tabular
(left) and time-series (right) tasks. Bold marks the best model per dataset.}
\label{tab:fidelity_heatmaps}
\end{table*}
 
\subsection{Statistical Fidelity}
 
Table~\ref{tab:fidelity_heatmaps} reports per-dataset Overall fidelity for all four generators on both modalities. Averaging across datasets, CoMed-TVAE is the most faithful generator on both modalities, attaining the highest mean Overall score (0.902 tabular, 0.887 time-series) and winning on the majority of individual datasets. On tabular data the other three generators are closely grouped (mean Overall 0.85–0.87), with CoMed-GaussianCopula (0.867) slightly ahead of the GAN-based CoMed-CTGAN (0.860) and CoMed-CopulaGAN (0.848). On time-series data CoMed-GaussianCopula is the clear runner-up (mean 0.880), while CoMed-CTGAN (0.833) and CoMed-CopulaGAN (0.831) trail. Table~\ref{tab:constraint_fidelity} isolates the effect of the clinical-validity layer: it improves Overall fidelity for the GAN-based CoMed-CTGAN and CoMed-CopulaGAN on nearly every affected task (up to $+0.098$ on cervical cancer), while the CoMed-TVAE and CoMed-GaussianCopula models change only marginally.


 
\subsection{Downstream Utility}
 
Table~\ref{tab:util} reports per-dataset downstream
utility (AUROC and AUPRC) for all four synthesizers. On tabular tasks the reference
generator CoMed-CTGAN preserves most of the real-data signal (mean AUC-ROC utility
$90.6\%$, median $89.3\%$), and the strongest generator CoMed-TVAE reaches $97.3\%$;
several datasets meet or exceed the real baseline (GBSG~$100.5\%$, Mammographic
Mass~$100.0\%$). Time-series tasks are harder and more generator-sensitive:
CoMed-CTGAN retains a mean AUC-ROC utility of $81.6\%$, whereas CoMed-TVAE still
retains ${\sim}95\%$. The gap widens sharply under AUC-PRC, which is sensitive to
minority-class recovery: for CoMed-CTGAN, tabular utility averages $86.4\%$ but
time-series falls to $64.0\%$, with rare-outcome ICU mortality tasks retaining only
${\sim}30$-$45\%$. Thus synthetic data transfers well for static tabular prediction,
but the weaker generators lose substantial minority-class signal on temporal, highly
imbalanced ICU tasks.

\begin{table*}[t]
\centering
\scriptsize
\setlength{\tabcolsep}{4pt}\renewcommand{\arraystretch}{1.15}
\captionsetup{font=footnotesize}
\resizebox{\textwidth}{!}{%
\begin{tabular}{@{}llccccc@{}}
\toprule
\textbf{Dataset} & \textbf{Task} & \textbf{Real} & \textbf{CoMed-CTGAN} & \textbf{CoMed-TVAE} & \textbf{CoMed-CopulaGAN} & \textbf{CoMed-GaussianCopula} \\
\midrule
\multicolumn{7}{@{}l}{\textit{Tabular}}\\
BC Coimbra & Breast Cancer & 0.777$\pm$0.124, 0.771$\pm$0.073 & 0.617$\pm$0.144, 0.671$\pm$0.094 & \textbf{0.731$\pm$0.066, 0.778$\pm$0.043} & 0.646$\pm$0.083, 0.752$\pm$0.051 & 0.710$\pm$0.109, 0.781$\pm$0.105 \\
CDC BRFSS & Diabetes & 0.819$\pm$0.012, 0.401$\pm$0.030 & 0.800$\pm$0.015, 0.377$\pm$0.023 & \textbf{0.809$\pm$0.011, 0.382$\pm$0.022} & 0.755$\pm$0.020, 0.309$\pm$0.024 & 0.758$\pm$0.038, 0.323$\pm$0.045 \\
GBSG & BC Survival & 0.663$\pm$0.022, 0.691$\pm$0.026 & 0.666$\pm$0.022, 0.710$\pm$0.011 & \textbf{0.676$\pm$0.007, 0.714$\pm$0.009} & 0.654$\pm$0.015, 0.699$\pm$0.006 & 0.661$\pm$0.007, 0.707$\pm$0.010 \\
METABRIC & BC Survival & 0.721$\pm$0.014, 0.786$\pm$0.018 & 0.667$\pm$0.027, 0.724$\pm$0.038 & \textbf{0.708$\pm$0.034, 0.781$\pm$0.028} & 0.690$\pm$0.011, 0.759$\pm$0.013 & 0.696$\pm$0.042, 0.764$\pm$0.032 \\
SUPPORT2 & Death & 0.848$\pm$0.005, 0.922$\pm$0.004 & 0.658$\pm$0.043, 0.795$\pm$0.028 & \textbf{0.827$\pm$0.005, 0.906$\pm$0.003} & 0.660$\pm$0.032, 0.802$\pm$0.025 & 0.764$\pm$0.008, 0.868$\pm$0.007 \\
SUPPORT2 & Mortality & 0.949$\pm$0.002, 0.893$\pm$0.006 & 0.847$\pm$0.036, 0.665$\pm$0.051 & \textbf{0.933$\pm$0.005, 0.852$\pm$0.010} & 0.824$\pm$0.037, 0.648$\pm$0.071 & 0.824$\pm$0.085, 0.660$\pm$0.114 \\
UCI Cervical Cancer & Cervical Cancer & 0.603$\pm$0.108, 0.125$\pm$0.051 & \textbf{0.533$\pm$0.053, 0.087$\pm$0.014} & 0.472$\pm$0.043, 0.091$\pm$0.037 & 0.493$\pm$0.084, 0.106$\pm$0.030 & 0.496$\pm$0.059, 0.075$\pm$0.013 \\
UCI CKD & CKD & 0.999$\pm$0.002, 0.998$\pm$0.003 & 0.989$\pm$0.004, 0.978$\pm$0.009 & \textbf{0.999$\pm$0.001, 0.999$\pm$0.001} & 0.990$\pm$0.005, 0.981$\pm$0.013 & 0.960$\pm$0.036, 0.934$\pm$0.059 \\
UCI Diabetes-130 & 30-day Readmit & 0.666$\pm$0.009, 0.216$\pm$0.009 & 0.513$\pm$0.009, 0.114$\pm$0.002 & 0.568$\pm$0.027, 0.156$\pm$0.021 & 0.548$\pm$0.023, 0.132$\pm$0.011 & \textbf{0.593$\pm$0.028, 0.154$\pm$0.023} \\
UCI Diabetes-130 & Any Readmit & 0.690$\pm$0.014, 0.653$\pm$0.013 & 0.614$\pm$0.008, 0.567$\pm$0.016 & \textbf{0.653$\pm$0.003, 0.609$\pm$0.012} & 0.611$\pm$0.013, 0.559$\pm$0.019 & 0.642$\pm$0.009, 0.601$\pm$0.016 \\
UCI Early Diabetes & Early Diabetes & 0.994$\pm$0.005, 0.997$\pm$0.003 & \textbf{0.988$\pm$0.002, 0.993$\pm$0.001} & 0.977$\pm$0.026, 0.987$\pm$0.015 & 0.963$\pm$0.006, 0.978$\pm$0.004 & 0.917$\pm$0.028, 0.953$\pm$0.016 \\
UCI Heart Failure & HF Mortality & 0.774$\pm$0.037, 0.599$\pm$0.062 & 0.691$\pm$0.047, 0.513$\pm$0.068 & \textbf{0.798$\pm$0.028, 0.693$\pm$0.071} & 0.768$\pm$0.028, 0.670$\pm$0.079 & 0.690$\pm$0.077, 0.571$\pm$0.076 \\
UCI Hepatitis & Hepatitis Mort. & 0.878$\pm$0.033, 0.972$\pm$0.009 & 0.867$\pm$0.050, 0.967$\pm$0.015 & \textbf{0.869$\pm$0.039, 0.969$\pm$0.009} & 0.702$\pm$0.155, 0.922$\pm$0.037 & 0.653$\pm$0.247, 0.886$\pm$0.104 \\
UCI ILPD & Liver Disease & 0.685$\pm$0.046, 0.436$\pm$0.075 & 0.669$\pm$0.028, 0.428$\pm$0.056 & \textbf{0.734$\pm$0.011, 0.472$\pm$0.013} & 0.603$\pm$0.032, 0.404$\pm$0.025 & 0.674$\pm$0.091, 0.435$\pm$0.087 \\
UCI Mamm.\ Mass & Malignancy & 0.886$\pm$0.016, 0.835$\pm$0.043 & 0.886$\pm$0.018, 0.860$\pm$0.024 & \textbf{0.902$\pm$0.008, 0.876$\pm$0.026} & 0.873$\pm$0.016, 0.834$\pm$0.030 & 0.700$\pm$0.096, 0.659$\pm$0.123 \\
UCI Statlog Heart & Heart Disease & 0.880$\pm$0.016, 0.842$\pm$0.033 & 0.852$\pm$0.032, 0.825$\pm$0.036 & \textbf{0.905$\pm$0.018, 0.888$\pm$0.024} & 0.846$\pm$0.022, 0.795$\pm$0.051 & 0.669$\pm$0.165, 0.639$\pm$0.147 \\
NHANES & Mortality & 0.920$\pm$0.005, 0.762$\pm$0.015 & 0.544$\pm$0.056, 0.224$\pm$0.038 & \textbf{0.893$\pm$0.029, 0.682$\pm$0.101} & 0.769$\pm$0.027, 0.409$\pm$0.040 & 0.868$\pm$0.049, 0.658$\pm$0.053 \\
\midrule
\multicolumn{7}{@{}l}{\textit{Time-series}}\\
MIMIC-III & COPD Mort. & 0.712$\pm$0.010, 0.343$\pm$0.027 & 0.412$\pm$0.026, 0.123$\pm$0.013 & \textbf{0.636$\pm$0.026, 0.227$\pm$0.026} & 0.445$\pm$0.045, 0.131$\pm$0.020 & 0.537$\pm$0.027, 0.190$\pm$0.027 \\
MIMIC-III & HF Mortality & 0.745$\pm$0.155, 0.413$\pm$0.170 & 0.480$\pm$0.109, 0.127$\pm$0.023 & \textbf{0.810$\pm$0.027, 0.453$\pm$0.030} & 0.461$\pm$0.059, 0.123$\pm$0.032 & 0.755$\pm$0.043, 0.336$\pm$0.045 \\
MIMIC-III & Hosp.\ Readmit & 0.600$\pm$0.017, 0.076$\pm$0.003 & 0.512$\pm$0.030, 0.058$\pm$0.006 & \textbf{0.596$\pm$0.022, 0.078$\pm$0.003} & 0.517$\pm$0.020, 0.057$\pm$0.005 & 0.513$\pm$0.054, 0.057$\pm$0.008 \\
MIMIC-III & Length-of-Stay & 0.757$\pm$0.014, 0.374$\pm$0.021 & 0.614$\pm$0.038, 0.238$\pm$0.019 & \textbf{0.699$\pm$0.024, 0.289$\pm$0.019} & 0.647$\pm$0.033, 0.256$\pm$0.016 & 0.612$\pm$0.063, 0.239$\pm$0.048 \\
MIMIC-III & In-Hosp. Mortality & 0.835$\pm$0.011, 0.467$\pm$0.019 & 0.690$\pm$0.019, 0.270$\pm$0.031 & \textbf{0.773$\pm$0.015, 0.358$\pm$0.015} & 0.665$\pm$0.015, 0.245$\pm$0.027 & 0.623$\pm$0.027, 0.210$\pm$0.027 \\
MIMIC-III & ICU Readmit & 0.601$\pm$0.016, 0.137$\pm$0.009 & 0.522$\pm$0.036, 0.107$\pm$0.013 & \textbf{0.592$\pm$0.014, 0.135$\pm$0.003} & 0.497$\pm$0.020, 0.099$\pm$0.005 & 0.512$\pm$0.036, 0.105$\pm$0.012 \\

MIMIC-IV & COPD Mort. & 0.827$\pm$0.014, 0.512$\pm$0.034 & 0.556$\pm$0.027, 0.220$\pm$0.031 & \textbf{0.770$\pm$0.006, 0.405$\pm$0.011} & 0.546$\pm$0.034, 0.228$\pm$0.032 & 0.557$\pm$0.074, 0.240$\pm$0.046 \\
MIMIC-IV & HF Mortality & 0.826$\pm$0.009, 0.474$\pm$0.013 & 0.555$\pm$0.061, 0.181$\pm$0.043 & \textbf{0.768$\pm$0.018, 0.372$\pm$0.016} & 0.632$\pm$0.058, 0.230$\pm$0.042 & 0.561$\pm$0.118, 0.209$\pm$0.083 \\
MIMIC-IV & Hosp.\ Readmit & 0.622$\pm$0.009, 0.359$\pm$0.014 & \textbf{0.519$\pm$0.009, 0.266$\pm$0.010} & 0.514$\pm$0.008, 0.269$\pm$0.012 & 0.502$\pm$0.011, 0.258$\pm$0.009 & 0.511$\pm$0.015, 0.265$\pm$0.011 \\
MIMIC-IV & Length-of-Stay & 0.828$\pm$0.008, 0.668$\pm$0.024 & \textbf{0.794$\pm$0.010, 0.606$\pm$0.017} & 0.766$\pm$0.013, 0.577$\pm$0.019 & 0.788$\pm$0.012, 0.605$\pm$0.018 & 0.750$\pm$0.026, 0.556$\pm$0.032 \\
MIMIC-IV & ICU Mort. & 0.895$\pm$0.004, 0.533$\pm$0.018 & 0.737$\pm$0.070, 0.239$\pm$0.051 & \textbf{0.861$\pm$0.005, 0.407$\pm$0.034} & 0.746$\pm$0.058, 0.252$\pm$0.040 & 0.675$\pm$0.119, 0.235$\pm$0.087 \\
MIMIC-IV & ICU Readmit & 0.644$\pm$0.022, 0.299$\pm$0.020 & 0.490$\pm$0.019, 0.191$\pm$0.010 & \textbf{0.609$\pm$0.058, 0.265$\pm$0.048} & 0.510$\pm$0.010, 0.196$\pm$0.005 & 0.518$\pm$0.023, 0.201$\pm$0.014 \\
eICU & COPD Mort. & 0.794$\pm$0.010, 0.365$\pm$0.009 & 0.556$\pm$0.088, 0.157$\pm$0.060 & \textbf{0.749$\pm$0.015, 0.280$\pm$0.019} & 0.596$\pm$0.057, 0.153$\pm$0.044 & 0.547$\pm$0.069, 0.154$\pm$0.038 \\
eICU & HF Mortality & 0.739$\pm$0.016, 0.327$\pm$0.015 & 0.612$\pm$0.042, 0.194$\pm$0.032 & \textbf{0.714$\pm$0.011, 0.285$\pm$0.026} & 0.598$\pm$0.036, 0.186$\pm$0.023 & 0.661$\pm$0.062, 0.246$\pm$0.055 \\
eICU & Length-of-Stay & 0.755$\pm$0.008, 0.284$\pm$0.012 & 0.692$\pm$0.033, 0.218$\pm$0.024 & \textbf{0.723$\pm$0.011, 0.239$\pm$0.011} & 0.694$\pm$0.033, 0.223$\pm$0.021 & 0.648$\pm$0.057, 0.199$\pm$0.044 \\
eICU & Mortality & 0.898$\pm$0.006, 0.438$\pm$0.028 & \textbf{0.870$\pm$0.016, 0.380$\pm$0.021} & 0.849$\pm$0.028, 0.359$\pm$0.051 & 0.870$\pm$0.012, 0.375$\pm$0.020 & 0.737$\pm$0.114, 0.222$\pm$0.109 \\
eICU & Readmit & 0.695$\pm$0.022, 0.358$\pm$0.042 & 0.595$\pm$0.022, 0.205$\pm$0.019 & \textbf{0.609$\pm$0.013, 0.213$\pm$0.007} & 0.576$\pm$0.021, 0.197$\pm$0.017 & 0.520$\pm$0.050, 0.173$\pm$0.029 \\
\midrule
\multicolumn{7}{@{}l}{\textit{Multiclass}}\\
SUPPORT2 & Functional & 0.822$\pm$0.010, 0.459$\pm$0.008 & 0.578$\pm$0.027, 0.274$\pm$0.018 & \textbf{0.771$\pm$0.013, 0.406$\pm$0.007} & 0.576$\pm$0.022, 0.303$\pm$0.007 & 0.600$\pm$0.029, 0.321$\pm$0.016 \\
UCI Dermatology & Dermatological Dis. & 0.996$\pm$0.003, 0.982$\pm$0.011 & 0.856$\pm$0.023, 0.593$\pm$0.039 & \textbf{0.997$\pm$0.002, 0.987$\pm$0.008} & 0.866$\pm$0.020, 0.576$\pm$0.039 & 0.740$\pm$0.040, 0.434$\pm$0.065 \\
UCI Diabetes-130 & Readmit (3-cls) & 0.669$\pm$0.015, 0.468$\pm$0.015 & 0.570$\pm$0.011, 0.384$\pm$0.007 & \textbf{0.614$\pm$0.007, 0.418$\pm$0.003} & 0.568$\pm$0.008, 0.380$\pm$0.005 & 0.607$\pm$0.016, 0.409$\pm$0.007 \\
\bottomrule
\end{tabular}%
}
\caption{Per-dataset downstream utility. Each cell reports \textbf{AUROC$\pm$std, AUPRC$\pm$std} (mean~$\pm$~std over the five classifiers; Real${=}$TRTR, generators${=}$TSTR). Best synthetic generator per row (by AUROC) in \textbf{bold}. Binary tasks use standard AUROC/AUPRC; the \textit{Multiclass} block uses macro one-vs-rest AUROC and macro AUPRC.}
\label{tab:util}
\end{table*}

\renewcommand{\topfraction}{0.95}
\renewcommand{\bottomfraction}{0.95}
\renewcommand{\textfraction}{0.03}
\renewcommand{\floatpagefraction}{0.90}

\setlength{\textfloatsep}{6pt plus 2pt minus 2pt}

\begin{figure}[!t]
\centering

\captionsetup{
    font=footnotesize,
    skip=3pt
}

\includegraphics[
    width=0.8\columnwidth
]{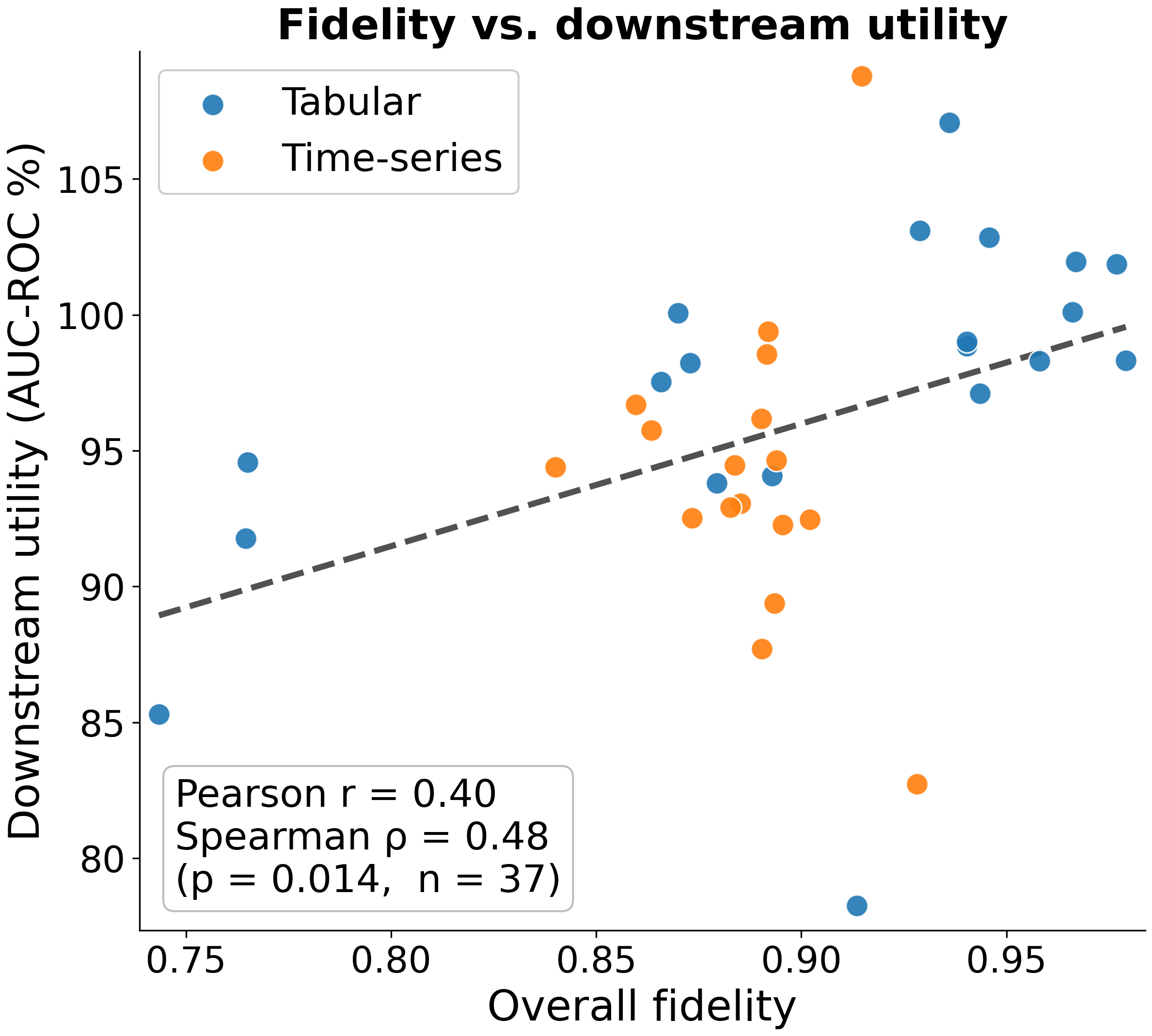}

\caption{Overall fidelity vs.\ downstream utility (CoMed-TVAE AUC-ROC utility)
across all $37$ dataset-task pairs. Overall fidelity is the strongest single
predictor of utility (Pearson $r=0.40$, Spearman $\rho=0.48$, $p=0.014$);
Correlation Stability is intermediate ($r=0.33$) and marginal Distribution
Stability weakest ($r=0.25$), while dataset size is not predictive
($r=-0.27$, $p=0.13$).}
\label{fig:fidelity_vs_utility}
\end{figure}

 The distributions in tabular utility are
higher and tighter (median near $93\%$ across classifier-dataset pairs), while
time-series utility is lower and more dispersed. The
choice of downstream classifier matters relatively little, so the data modality, not
the classifier, drives utility.
 Notably, on the imbalanced ICU tasks a synthetic dataset can retain acceptable AUROC while its AUPRC collapses, so AUROC alone overstates usefulness for rare-outcome prediction.
 

\subsection{Does Fidelity Predict Utility?}

Finally we relate the two axes across all
$37$ dataset-task pairs: higher fidelity is
associated with higher utility, but the
strength depends on the property. For the reference generator CoMed-CTGAN, Overall fidelity is the best single predictor
($r=0.67$, $\rho=0.74$), Correlation Stability is intermediate ($r=0.58$), and
marginal Distribution Stability is weakest ($r=0.45$). Dataset size shows no
significant relationship ($r=-0.20$, $p=0.26$). The same ranking of predictors
holds for the strongest generator, CoMed-TVAE
(Figure~\ref{fig:fidelity_vs_utility}), although the correlation is attenuated
(Overall $r=0.40$, $\rho=0.48$) because its near-ceiling utility leaves little
variance for fidelity to explain. These correlations are positive but
imperfect, reinforcing the benchmark's premise: fidelity is a useful but incomplete
proxy, and downstream utility must be measured directly.

\section{Conclusion}
 
We presented CoMedBench, a reproducible benchmark that evaluates synthetic clinical
data under a common clinical-validity framework along two axes-statistical
fidelity and downstream task utility-using a single configuration-driven engine
over three intensive-care databases and a suite of public clinical and tabular
datasets. Across four generators, no single one dominates: CoMed-TVAE is the most
faithful overall, while the Gaussian-copula model CoMed-GC is competitive or best on
several small or highly imbalanced cohorts.
 
Our central finding concerns usefulness rather than appearance. Synthetic tabular
data preserves most downstream signal (mean AUC-ROC utility $90.6\%$ for CoMed-CTGAN,
up to $97.3\%$ for CoMed-TVAE), and several datasets match the real-data upper bound.
Temporal ICU tasks are harder and more generator-sensitive: under CoMed-CTGAN utility
falls to $81.6\%$ under AUC-ROC and to $64.0\%$ under the imbalance-sensitive
AUC-PRC, though CoMed-TVAE still retains ${\sim}95\%$ under AUC-ROC. Relating the two
axes, fidelity-especially the preservation of pairwise correlation structure-is
positively but imperfectly correlated with utility (for CoMed-CTGAN, Overall
$r=0.67$, $\rho=0.74$), while dataset size is not. Statistical fidelity is therefore
a useful but insufficient proxy, and downstream utility must be evaluated directly.

The practical message is that synthetic data is already a credible substitute for
static tabular prototyping and benchmarking, while rare-event imbalance remains the
hardest case. Because our pipeline summarizes each ICU stay into a per-stay feature
vector, claims about fine-grained temporal dynamics are out of scope; this motivates
comparison against sequence-native generators. Integrating privacy tests would
complete the fidelity-utility-privacy picture needed for clinical deployment.



\bibliography{aaai2027.bib}

@article{johnson2016mimic,
  title={MIMIC-III, a Freely Accessible Critical Care Database},
  author={Johnson, Alistair E. W. and Pollard, Tom J. and Shen, Lu and Lehman, Li-wei H. and Feng, Mengling and Ghassemi, Mohammad and Moody, Benjamin and Szolovits, Peter and Celi, Leo Anthony and Mark, Roger G.},
  journal={Scientific Data},
  volume={3},
  pages={160035},
  year={2016}
}

@article{johnson2023mimiciv,
  title={MIMIC-IV, a Freely Accessible Electronic Health Record Dataset},
  author={Johnson, Alistair E. W. and Bulgarelli, Lucas and Shen, Lu and Gayles, Alvin and Shammout, Ayad and Horng, Steven and Pollard, Tom J. and Hao, Sicheng and Moody, Benjamin and Gow, Brian and Lehman, Li-wei H. and Celi, Leo Anthony and Mark, Roger G.},
  journal={Scientific Data},
  volume={10},
  pages={1},
  year={2023}
}

@article{pollard2018eicu,
  title={The eICU Collaborative Research Database, a Freely Available Multi-Center Database for Critical Care Research},
  author={Pollard, Tom J. and Johnson, Alistair E. W. and Raffa, Jesse D. and Celi, Leo Anthony and Mark, Roger G. and Badawi, Omar},
  journal={Scientific Data},
  volume={5},
  pages={180178},
  year={2018}
}

@inproceedings{xu2019ctgan,
  title={Modeling Tabular Data Using Conditional GAN},
  author={Xu, Lei and Skoularidou, Maria and Cuesta-Infante, Alfredo and Veeramachaneni, Kalyan},
  booktitle={Advances in Neural Information Processing Systems},
  year={2019}
}

@inproceedings{patki2016sdv,
  title={The Synthetic Data Vault},
  author={Patki, Neha and Wedge, Roy and Veeramachaneni, Kalyan},
  booktitle={IEEE International Conference on Data Science and Advanced Analytics},
  pages={399--410},
  year={2016}
}

@article{che2018grud,
  title={Recurrent Neural Networks for Multivariate Time Series with Missing Values},
  author={Che, Zhengping and Purushotham, Sanjay and Cho, Kyunghyun and Sontag, David and Liu, Yan},
  journal={Scientific Reports},
  volume={8},
  pages={6085},
  year={2018}
}

@article{harutyunyan2019multitask,
  title={Multitask Learning and Benchmarking with Clinical Time Series Data},
  author={Harutyunyan, Hrayr and Khachatrian, Hrant and Kale, David C. and Ver Steeg, Greg and Galstyan, Aram},
  journal={Scientific Data},
  volume={6},
  pages={96},
  year={2019}
}

@misc{gupta2022mimicivpipeline,
  title={MIMIC-IV Data Pipeline for Clinical Prediction Tasks},
  author={Gupta, Mayank and Gallamoza, Benjamin and Cutrona, Nicholas and Dhakal, Piyush and Poulain, Raphael and Beheshti, Rahmatollah},
  year={2022},
  eprint={2204.13841},
  archivePrefix={arXiv},
  primaryClass={cs.LG}
}

@inproceedings{lee2022dpflows,
  title={Differentially Private Normalizing Flows for Synthetic Tabular Data Generation},
  author={Lee, Jaewoo and Kim, Minjung and Jeong, Yonghyun and Ro, Youngmin},
  booktitle={Proceedings of the AAAI Conference on Artificial Intelligence},
  year={2022}
}

@ARTICLE{11186191,
  author={Pias, Tanmoy Sarkar and Su, Yiqi and Tang, Xuxin and Wang, Haohui and Faghani, Shahriar and Yao, Danfeng},
  journal={IEEE Journal of Biomedical and Health Informatics}, 
  title={Enhancing Fairness and Accuracy in Diagnosing Type 2 Diabetes in Young Adult Population}, 
  year={2026},
  volume={30},
  number={4},
  pages={3321-3330},
  doi={10.1109/JBHI.2025.3616312}}

@article{Pias2025,
  author  = {Pias, Tanmoy Sarkar and Afrose, Sharmin and Tuli, Moon Das and
             Trisha, Ipsita Hamid and Deng, Xinwei and Nemeroff, Charles B. and
             Yao, Danfeng Daphne},
  title   = {Low responsiveness of machine learning models to critical or
             deteriorating health conditions},
  journal = {Communications Medicine},
  year    = {2025},
  volume  = {5},
  number  = {1},
  pages   = {62},
  doi     = {10.1038/s43856-025-00775-0},
  url     = {https://doi.org/10.1038/s43856-025-00775-0}
}

@article{habehh2021machine,
  title={Machine learning in healthcare},
  author={Habehh, Hafsa and Gohel, Suril},
  journal={Current genomics},
  volume={22},
  number={4},
  pages={291--300},
  year={2021},
  publisher={Bentham Science Publishers direct}
}

@misc{adisa2026integratedframeworkexplainablefair,
      title={An Integrated Framework for Explainable, Fair, and Observable Hospital Readmission Prediction: Development and Validation on MIMIC-IV}, 
      author={Isaac Tosin Adisa},
      year={2026},
      eprint={2604.22535},
      archivePrefix={arXiv},
      primaryClass={cs.LG},
      url={https://arxiv.org/abs/2604.22535}, 
}

@article{li2021prediction,
  title={Prediction model of in-hospital mortality in intensive care unit patients with heart failure: machine learning-based, retrospective analysis of the MIMIC-III database},
  author={Li, Fuhai and Xin, Hui and Zhang, Jidong and Fu, Mingqiang and Zhou, Jingmin and Lian, Zhexun},
  journal={BMJ open},
  volume={11},
  number={7},
  pages={e044779},
  year={2021},
  publisher={British Medical Journal Publishing Group}
}

@article{tang2020democratizing,
  title={Democratizing EHR analyses with FIDDLE: a flexible data-driven preprocessing pipeline for structured clinical data},
  author={Tang, Shengpu and Davarmanesh, Parmida and Song, Yanmeng and Koutra, Danai and Sjoding, Michael W and Wiens, Jenna},
  journal={Journal of the American Medical Informatics Association},
  volume={27},
  number={12},
  pages={1921--1934},
  year={2020},
  publisher={Oxford University Press}
}

@article{chicco2020machine,
  title={Machine learning can predict survival of patients with heart failure from serum creatinine and ejection fraction alone},
  author={Chicco, Davide and Jurman, Giuseppe},
  journal={BMC medical informatics and decision making},
  volume={20},
  number={1},
  pages={16},
  year={2020},
  publisher={Springer}
}

@article{xie2019building,
  title={Building risk prediction models for type 2 diabetes using machine learning techniques},
  author={Xie, Zidian and Nikolayeva, Olga and Luo, Jiebo and Li, Dongmei},
  journal={Preventing chronic disease},
  volume={16},
  pages={E130},
  year={2019}
}

@article{strack2014impact,
  title={Impact of HbA1c measurement on hospital readmission rates: analysis of 70,000 clinical database patient records},
  author={Strack, Beata and DeShazo, Jonathan P and Gennings, Chris and Olmo, Juan L and Ventura, Sebastian and Cios, Krzysztof J and Clore, John N},
  journal={BioMed research international},
  volume={2014},
  number={1},
  pages={781670},
  year={2014},
  publisher={Wiley Online Library}
}

@article{ramana2011critical,
  title={A critical study of selected classification algorithms for liver disease diagnosis},
  author={Ramana, Bendi Venkata and Babu, M Surendra Prasad and Venkateswarlu, NB and others},
  journal={International Journal of Database Management Systems},
  volume={3},
  number={2},
  pages={101--114},
  year={2011},
  publisher={Academy \& Industry Research Collaboration Center(AIRCC)}
}

@article{patricio2018using,
  title={Using Resistin, glucose, age and BMI to predict the presence of breast cancer},
  author={Patr{\'\i}cio, Miguel and Pereira, Jos{\'e} and Cris{\'o}stomo, Joana and Matafome, Paulo and Gomes, Manuel and Sei{\c{c}}a, Raquel and Caramelo, Francisco},
  journal={BMC cancer},
  volume={18},
  number={1},
  pages={29},
  year={2018},
  publisher={Springer}
}

@article{katzman2018deepsurv,
  title={DeepSurv: personalized treatment recommender system using a Cox proportional hazards deep neural network},
  author={Katzman, Jared L and Shaham, Uri and Cloninger, Alexander and Bates, Jonathan and Jiang, Tingting and Kluger, Yuval},
  journal={BMC medical research methodology},
  volume={18},
  number={1},
  pages={24},
  year={2018},
  publisher={Springer}
}

@article{qiu2022interpretable,
  title={Interpretable machine learning prediction of all-cause mortality},
  author={Qiu, Wei and Chen, Hugh and Dincer, Ayse Berceste and Lundberg, Scott and Kaeberlein, Matt and Lee, Su-In},
  journal={Communications medicine},
  volume={2},
  number={1},
  pages={125},
  year={2022},
  publisher={Nature Publishing Group UK London}
}

@article{peng2022interpretable,
  title={Interpretable machine learning for 28-day all-cause in-hospital mortality prediction in critically ill patients with heart failure combined with hypertension: a retrospective cohort study based on medical information mart for intensive care database-IV and eICU databases},
  author={Peng, Shengxian and Huang, Jian and Liu, Xiaozhu and Deng, Jiewen and Sun, Chenyu and Tang, Juan and Chen, Huaqiao and Cao, Wenzhai and Wang, Wei and Duan, Xiangjie and others},
  journal={Frontiers in cardiovascular medicine},
  volume={9},
  pages={994359},
  year={2022},
  publisher={Frontiers Media SA}
}

@article{bui2024benchmarking,
  title={Benchmarking with MIMIC-IV, an irregular, spare clinical time series dataset},
  author={Bui, Hung and Warrier, Harikrishna and Gupta, Yogesh},
  journal={arXiv preprint arXiv:2401.15290},
  year={2024}
}

@misc{kakadiaris2023evaluatingfairnessmimicivdataset,
      title={Evaluating the Fairness of the MIMIC-IV Dataset and a Baseline Algorithm: Application to the ICU Length of Stay Prediction}, 
      author={Alexandra Kakadiaris},
      year={2023},
      eprint={2401.00902},
      archivePrefix={arXiv},
      primaryClass={cs.LG},
      url={https://arxiv.org/abs/2401.00902}, 
}

@article{lin2019analysis,
  title={Analysis and prediction of unplanned intensive care unit readmission using recurrent neural networks with long short-term memory},
  author={Lin, Yu-Wei and Zhou, Yuqian and Faghri, Faraz and Shaw, Michael J and Campbell, Roy H},
  journal={PloS one},
  volume={14},
  number={7},
  pages={e0218942},
  year={2019},
  publisher={Public Library of Science San Francisco, CA USA}
}

@inproceedings{gupta2022extensive,
  title={An extensive data processing pipeline for mimic-iv},
  author={Gupta, Mehak and Gallamoza, Brennan and Cutrona, Nicolas and Dhakal, Pranjal and Poulain, Raphael and Beheshti, Rahmatollah},
  booktitle={Machine learning for health},
  pages={311--325},
  year={2022},
  organization={PMLR}
}

@article{Tucker2020,
  author  = {Tucker, Allan and Wang, Zhenchen and Rotalinti, Ylenia and Myles, Puja},
  title   = {Generating High-Fidelity Synthetic Patient Data for Assessing Machine Learning Healthcare Software},
  journal = {npj Digital Medicine},
  year    = {2020},
  volume  = {3},
  number  = {1},
  pages   = {147},
  doi     = {10.1038/s41746-020-00353-9},
  url     = {https://doi.org/10.1038/s41746-020-00353-9},
  issn    = {2398-6352}
}

@article{Chen2021,
  author  = {Chen, Richard J. and Lu, Ming Y. and Chen, Tiffany Y. and Williamson, Drew F. K. and Mahmood, Faisal},
  title   = {Synthetic Data in Machine Learning for Medicine and Healthcare},
  journal = {Nature Biomedical Engineering},
  year    = {2021},
  volume  = {5},
  number  = {6},
  pages   = {493--497},
  doi     = {10.1038/s41551-021-00751-8},
  url     = {https://doi.org/10.1038/s41551-021-00751-8},
  issn    = {2157-846X}
}

\end{document}


\onecolumn
\thispagestyle{plain}
\begin{center}
{\LARGE\bfseries CoMedBench: A Multi-Source Benchmark of Synthetic Medical Data Fidelity and Downstream Utility}\\[5pt]
{\large\bfseries Supplementary Material}\\[10pt]
\end{center}

\noindent
This supplement provides the complete, non-aggregated benchmark results that the main
paper summarizes for space. Results are reported at full granularity: one row per
dataset$\times$generator for statistical fidelity, and one row per
dataset$\times$task$\times$classifier for downstream utility.\par\smallskip

\begingroup
\footnotesize\setlength{\tabcolsep}{4pt}\renewcommand{\arraystretch}{0.95}
\begin{longtable}{@{}l l p{150pt}|l p{150pt}@{}}
\caption{Abbreviation key used in the result tables.}\label{tab:abbr-key}\\
\toprule
\textbf{Group} & \textbf{Abbr.} & \textbf{Meaning} & \textbf{Abbr.} & \textbf{Meaning} \\
\midrule\endfirsthead
\multicolumn{5}{@{}l}{\emph{\tablename~\thetable\ (continued)}}\\
\toprule
\textbf{Group} & \textbf{Abbr.} & \textbf{Meaning} & \textbf{Abbr.} & \textbf{Meaning} \\
\midrule\endhead
\midrule\multicolumn{5}{r@{}}{\emph{continued on next page}}\\\endfoot
\bottomrule\endlastfoot
Generators & CTGAN & CoMed-CTGAN & TVAE & CoMed-TVAE \\*
 & CG & CoMed-CopulaGAN & GC & CoMed-GaussianCopula \\
Classifiers & LR & Logistic Regression & RF & Random Forest \\*
 & GB & Gradient Boosting & XGB & XGBoost \\*
 & MLP & multilayer perceptron & & \\
Datasets & BCC & Breast Cancer Coimbra & UCI-CC & UCI Cervical Cancer \\*
 & UCI-CKD & UCI Chronic Kidney Disease & UCI-Derm & UCI Dermatology \\*
 & UCI-D130 & UCI Diabetes 130-US Hospitals & UCI-ESD & UCI Early-Stage Diabetes \\*
 & UCI-HF & UCI Heart Failure & UCI-ILP & UCI Indian Liver Patient \\*
 & UCI-MM & UCI Mammographic Mass & UCI-SH & UCI Statlog Heart \\
Tasks & BC & Breast Cancer & BCS & Breast Cancer Survival \\*
 & CC & Cervical Cancer & CKD & Chronic Kidney Disease \\*
 & DDC & Dermatological Disease Classification & 30d-R & 30-day Readmission \\*
 & Any-R & Any-Readmission & R-3 & Readmission (3-class) \\*
 & EDR & Early Diabetes Risk & HFM & Heart Failure Mortality \\*
 & Hep-M & Hepatitis Mortality & LD & Liver Disease \\*
 & MM & Mass Malignancy & HD & Heart Disease \\*
 & COPD-M & COPD Mortality & Hosp-R & Hospital Readmission \\*
 & LOS & Length-of-Stay & ICU-R & ICU Readmission \\*
 & Hosp-M & Hospital Mortality & ICU-M & ICU Mortality \\*
 & FO & Functional Outcome & & \\
Metrics & TRTR & train-real/test-real & TSTR & train-synthetic/test-real \\*
 & Ret. & TSTR/TRTR & -- & unavailable run \\
\end{longtable}
\endgroup
\vspace{-2pt}

\noindent Overall Quality is the mean of Distribution Stability and Correlation Stability.
All numerical values match the experiments reported in the main paper.

\section{Datasets and Prediction Tasks}
\begingroup
\footnotesize\setlength{\tabcolsep}{4pt}\renewcommand{\arraystretch}{0.92}
\begin{longtable}{@{}l l l r l@{}}
\caption{Tabular datasets and prediction tasks.}\label{tab:supp-data-tab}\\
\toprule
\textbf{Dataset} & \textbf{Task} & \textbf{Task type} & \textbf{Rows} & \textbf{\#num / \#cat} \\
\midrule\endfirsthead
\multicolumn{5}{@{}l}{\emph{\tablename~\thetable\ (continued)}}\\
\toprule
\textbf{Dataset} & \textbf{Task} & \textbf{Task type} & \textbf{Rows} & \textbf{\#num / \#cat} \\
\midrule\endhead
\midrule\multicolumn{5}{r@{}}{\emph{continued on next page}}\\\endfoot
\bottomrule\endlastfoot
Breast Cancer Coimbra (BCC) & Breast Cancer & Binary & 116 & 9 / 0 \\
CDC BRFSS & Diabetes & Binary & 253,680 & 3 / 18 \\
GBSG (pycox) & Breast Cancer Survival & Binary & 2,232 & 4 / 3 \\
METABRIC (pycox) & Breast Cancer Survival & Binary & 1,904 & 5 / 4 \\
SUPPORT2 & Death & Binary & 9,105 & 24 / 13 \\*
SUPPORT2 & Functional Outcome & Multi-class (5) & 7,705 & 24 / 13 \\*
SUPPORT2 & Mortality & Binary & 9,105 & 24 / 13 \\
UCI Cervical Cancer & Cervical Cancer & Binary & 858 & 6 / 22 \\
UCI Chronic Kidney Disease & Chronic Kidney Disease & Binary & 400 & 11 / 13 \\
UCI Dermatology & Dermatological Disease Classification & Multi-class (6) & 366 & 1 / 33 \\
UCI Diabetes 130-US Hospitals & 30-day Readmission & Binary & 101,766 & 8 / 36 \\*
UCI Diabetes 130-US Hospitals & Any-Readmission & Binary & 101,766 & 8 / 36 \\*
UCI Diabetes 130-US Hospitals & Readmission (3-class) & Multi-class (3) & 101,766 & 8 / 36 \\
UCI Early-Stage Diabetes & Early Diabetes Risk & Binary & 520 & 1 / 15 \\
UCI Heart Failure & Heart Failure Mortality & Binary & 299 & 6 / 5 \\
UCI Hepatitis & Hepatitis Mortality & Binary & 155 & 6 / 13 \\
UCI Indian Liver Patient & Liver Disease & Binary & 583 & 9 / 1 \\
UCI Mammographic Mass & Mass Malignancy & Binary & 961 & 1 / 4 \\
UCI Statlog Heart & Heart Disease & Binary & 270 & 5 / 8 \\
NHANES & Mortality & Binary & 33,654 & 70 / 81 \\
\end{longtable}
\endgroup
\vspace{-2pt}

\begingroup
\footnotesize\setlength{\tabcolsep}{4pt}\renewcommand{\arraystretch}{0.92}
\begin{longtable}{@{}l l l r l@{}}
\caption{Time-series (ICU) datasets and prediction tasks.}\label{tab:supp-data-ts}\\
\toprule
\textbf{Dataset} & \textbf{Task} & \textbf{Task type} & \textbf{Rows} & \textbf{\#num / \#cat} \\
\midrule\endfirsthead
\multicolumn{5}{@{}l}{\emph{\tablename~\thetable\ (continued)}}\\
\toprule
\textbf{Dataset} & \textbf{Task} & \textbf{Task type} & \textbf{Rows} & \textbf{\#num / \#cat} \\
\midrule\endhead
\midrule\multicolumn{5}{r@{}}{\emph{continued on next page}}\\\endfoot
\bottomrule\endlastfoot
MIMIC-III & COPD Mortality & Binary & 3,285 & 109 / 2 \\*
MIMIC-III & Heart Failure Mortality & Binary & 1,176 & 38 / 10 \\*
MIMIC-III & Hospital Readmission & Binary & 29,000 & 109 / 2 \\*
MIMIC-III & Length-of-Stay & Binary & 32,551 & 109 / 2 \\*
MIMIC-III & ICU Readmission & Binary & 29,000 & 109 / 2 \\*
MIMIC-III & Hospital Mortality & Binary & 14,681 & 168 / 3 \\
MIMIC-IV & COPD Mortality & Binary & 6,385 & 126 / 3 \\*
MIMIC-IV & Heart Failure Mortality & Binary & 13,838 & 130 / 3 \\*
MIMIC-IV & Hospital Readmission & Binary & 59,646 & 34 / 249 \\*
MIMIC-IV & Length-of-Stay & Binary & 74,829 & 39 / 4 \\*
MIMIC-IV & ICU Mortality & Binary & 51,838 & 126 / 3 \\*
MIMIC-IV & ICU Readmission & Binary & 9,648 & 62 / 280 \\
eICU & COPD Mortality & Binary & 7,735 & 35 / 9 \\*
eICU & Heart Failure Mortality & Binary & 10,377 & 35 / 9 \\*
eICU & Length-of-Stay & Binary & 91,852 & 34 / 9 \\*
eICU & Mortality & Binary & 111,390 & 34 / 28 \\*
eICU & Readmission & Binary & 104,639 & 104 / 9 \\
\end{longtable}
\endgroup
\par\medskip

\Needspace{12\baselineskip}
\section{Statistical Fidelity: Full Per-Dataset Results}
\begingroup
\scriptsize\setlength{\tabcolsep}{10.8pt}\renewcommand{\arraystretch}{1.0}
\Needspace{11\baselineskip}
\begin{longtable}{@{}l l l r r r r r@{}}
\caption{Per-dataset tabular statistical fidelity (SDV) for the four generators. Best Overall Quality per dataset in \textbf{bold}. Dist. $=$ Distribution Stability; Corr. $=$ Correlation Stability; Valid. $=$ Data Validity; Struct. $=$ Data Structure. Abbreviations: see Table~\ref{tab:abbr-key}.}\label{tab:supp-fid-tab}\\
\toprule
\textbf{Dataset} & \textbf{Task} & \textbf{Gen.} & \textbf{Dist.} & \textbf{Corr.} & \textbf{Overall} & \textbf{Valid.} & \textbf{Struct.} \\
\midrule\endfirsthead
\multicolumn{8}{@{}l}{\emph{\tablename~\thetable\ (continued)}}\\
\toprule
\textbf{Dataset} & \textbf{Task} & \textbf{Gen.} & \textbf{Dist.} & \textbf{Corr.} & \textbf{Overall} & \textbf{Valid.} & \textbf{Struct.} \\
\midrule\endhead
\midrule\multicolumn{8}{r@{}}{\emph{continued on next page}}\\\endfoot
\bottomrule\endlastfoot
BCC & BC & CTGAN & 0.828 & 0.661 & 0.744 & 1.000 & 1.000 \\*
BCC & BC & TVAE & 0.860 & 0.925 & 0.893 & 1.000 & 1.000 \\*
BCC & BC & CG & 0.812 & 0.708 & 0.760 & 1.000 & 1.000 \\*
BCC & BC & GC & 0.828 & 0.963 & \textbf{0.895} & 1.000 & 1.000 \\
CDC BRFSS & Diabetes & CTGAN & 0.953 & 0.988 & \textbf{0.970} & 1.000 & 1.000 \\*
CDC BRFSS & Diabetes & TVAE & 0.953 & 0.928 & 0.940 & 1.000 & 1.000 \\*
CDC BRFSS & Diabetes & CG & 0.905 & 0.975 & 0.940 & 1.000 & 1.000 \\*
CDC BRFSS & Diabetes & GC & 0.934 & 0.765 & 0.849 & 1.000 & 1.000 \\
\end{longtable}
\endgroup
\par\medskip

\begingroup
\scriptsize\setlength{\tabcolsep}{10.8pt}\renewcommand{\arraystretch}{1.0}
\Needspace{11\baselineskip}
\begin{longtable}{@{}l l l r r r r r@{}}
\caption{Per-dataset time-series statistical fidelity (SDV) for the four generators. Best Overall Quality per dataset in \textbf{bold}. Dist. $=$ Distribution Stability; Corr. $=$ Correlation Stability; Valid. $=$ Data Validity; Struct. $=$ Data Structure. Abbreviations: see Table~\ref{tab:abbr-key}.}\label{tab:supp-fid-ts}\\
\toprule
\textbf{Dataset} & \textbf{Task} & \textbf{Gen.} & \textbf{Dist.} & \textbf{Corr.} & \textbf{Overall} & \textbf{Valid.} & \textbf{Struct.} \\
\midrule\endfirsthead
\multicolumn{8}{@{}l}{\emph{\tablename~\thetable\ (continued)}}\\
\toprule
\textbf{Dataset} & \textbf{Task} & \textbf{Gen.} & \textbf{Dist.} & \textbf{Corr.} & \textbf{Overall} & \textbf{Valid.} & \textbf{Struct.} \\
\midrule\endhead
\midrule\multicolumn{8}{r@{}}{\emph{continued on next page}}\\\endfoot
\bottomrule\endlastfoot
MIMIC-III & COPD-M & CTGAN & 0.860 & 0.733 & 0.796 & 1.000 & 1.000 \\*
MIMIC-III & COPD-M & TVAE & 0.876 & 0.910 & 0.893 & 1.000 & 1.000 \\*
MIMIC-III & COPD-M & CG & 0.849 & 0.732 & 0.791 & 1.000 & 1.000 \\*
MIMIC-III & COPD-M & GC & 0.846 & 0.963 & \textbf{0.904} & 1.000 & 1.000 \\
MIMIC-III & HFM & CTGAN & 0.821 & 0.688 & 0.754 & 1.000 & 1.000 \\*
MIMIC-III & HFM & TVAE & 0.917 & 0.912 & 0.915 & 1.000 & 1.000 \\*
MIMIC-III & HFM & CG & 0.807 & 0.691 & 0.749 & 1.000 & 1.000 \\*
MIMIC-III & HFM & GC & 0.923 & 0.986 & \textbf{0.955} & 1.000 & 1.000 \\
\end{longtable}
\endgroup
\par\medskip

\Needspace{12\baselineskip}
\section{Downstream Utility: Tabular (Per-Classifier)}
\begingroup
\scriptsize\setlength{\tabcolsep}{2pt}\renewcommand{\arraystretch}{1.0}
\Needspace{11\baselineskip}
\begin{longtable}{@{}l l l c c c c c c@{}}
\caption{Per-dataset, per-classifier tabular downstream utility, AUROC (with the clinical-validity layer). \utilitycolumnkey}\label{tab:supp-util-tab-auroc}\\
\toprule
\textbf{Dataset} & \textbf{Task} & \textbf{Classifier} & \textbf{Paper} & \textbf{Real} & \textbf{CTGAN} & \textbf{TVAE} & \textbf{CG} & \textbf{GC} \\
\midrule\endfirsthead
\multicolumn{9}{@{}l}{\emph{\tablename~\thetable\ (continued)}}\\
\toprule
\textbf{Dataset} & \textbf{Task} & \textbf{Classifier} & \textbf{Paper} & \textbf{Real} & \textbf{CTGAN} & \textbf{TVAE} & \textbf{CG} & \textbf{GC} \\
\midrule\endhead
\midrule\multicolumn{9}{r@{}}{\emph{continued on next page}}\\\endfoot
\bottomrule\endlastfoot
BCC & BC & LR & -- & 0.776 & 0.762 (98.2\%) & 0.678 (87.4\%) & 0.769 (99.1\%) & 0.615 (79.3\%) \\*
BCC & BC & RF & -- & 0.846 & 0.647 (76.5\%) & 0.783 (92.6\%) & 0.573 (67.8\%) & 0.787 (93.0\%) \\*
BCC & BC & GB & -- & 0.881 & 0.622 (70.6\%) & 0.776 (88.1\%) & 0.587 (66.7\%) & 0.755 (85.7\%) \\*
BCC & BC & XGB & -- & 0.818 & 0.678 (82.9\%) & 0.776 (94.9\%) & 0.608 (74.4\%) & 0.818 (100.0\%) \\*
BCC & BC & MLP & -- & 0.566 & 0.378 (66.7\%) & 0.643 (113.6\%) & 0.692 (122.2\%) & 0.573 (101.2\%) \\
CDC BRFSS & Diabetes & LR & -- & 0.820 & 0.812 (99.0\%) & 0.815 (99.5\%) & 0.768 (93.7\%) & 0.799 (97.5\%) \\*
CDC BRFSS & Diabetes & RF & -- & 0.798 & 0.775 (97.1\%) & 0.791 (99.1\%) & 0.739 (92.6\%) & 0.715 (89.6\%) \\*
CDC BRFSS & Diabetes & GB & -- & 0.825 & 0.809 (98.0\%) & 0.817 (99.0\%) & 0.779 (94.4\%) & 0.785 (95.1\%) \\
\end{longtable}
\endgroup
\par\medskip

\begingroup
\scriptsize\setlength{\tabcolsep}{2pt}\renewcommand{\arraystretch}{1.0}
\Needspace{11\baselineskip}
\begin{longtable}{@{}l l l c c c c c c@{}}
\caption{Per-dataset, per-classifier tabular downstream utility, AUPRC (with the clinical-validity layer). \utilitycolumnkey}\label{tab:supp-util-tab-auprc}\\
\toprule
\textbf{Dataset} & \textbf{Task} & \textbf{Classifier} & \textbf{Paper} & \textbf{Real} & \textbf{CTGAN} & \textbf{TVAE} & \textbf{CG} & \textbf{GC} \\
\midrule\endfirsthead
\multicolumn{9}{@{}l}{\emph{\tablename~\thetable\ (continued)}}\\
\toprule
\textbf{Dataset} & \textbf{Task} & \textbf{Classifier} & \textbf{Paper} & \textbf{Real} & \textbf{CTGAN} & \textbf{TVAE} & \textbf{CG} & \textbf{GC} \\
\midrule\endhead
\midrule\multicolumn{9}{r@{}}{\emph{continued on next page}}\\\endfoot
\bottomrule\endlastfoot
BCC & BC & LR & -- & 0.831 & 0.763 (91.9\%) & 0.743 (89.4\%) & 0.827 (99.5\%) & 0.705 (84.8\%) \\*
BCC & BC & RF & -- & 0.811 & 0.664 (81.9\%) & 0.792 (97.6\%) & 0.725 (89.4\%) & 0.844 (104.0\%) \\*
BCC & BC & GB & -- & 0.807 & 0.667 (82.6\%) & 0.827 (102.4\%) & 0.750 (92.9\%) & 0.839 (104.0\%) \\*
BCC & BC & XGB & -- & 0.754 & 0.740 (98.2\%) & 0.803 (106.6\%) & 0.766 (101.6\%) & 0.881 (116.9\%) \\*
BCC & BC & MLP & -- & 0.651 & 0.523 (80.2\%) & 0.724 (111.0\%) & 0.690 (105.9\%) & 0.636 (97.6\%) \\
CDC BRFSS & Diabetes & LR & -- & 0.393 & 0.381 (97.1\%) & 0.388 (98.7\%) & 0.326 (83.0\%) & 0.367 (93.4\%) \\*
CDC BRFSS & Diabetes & RF & -- & 0.351 & 0.337 (95.8\%) & 0.345 (98.3\%) & 0.274 (78.0\%) & 0.253 (71.9\%) \\*
CDC BRFSS & Diabetes & GB & -- & 0.420 & 0.392 (93.3\%) & 0.403 (96.0\%) & 0.336 (80.1\%) & 0.341 (81.2\%) \\
\end{longtable}
\endgroup
\par\medskip

\Needspace{12\baselineskip}
\section{Downstream Utility: Time-Series (Per-Classifier)}
\begingroup
\scriptsize\setlength{\tabcolsep}{2pt}\renewcommand{\arraystretch}{1.0}
\Needspace{11\baselineskip}
\begin{longtable}{@{}l l l c c c c c c@{}}
\caption{Per-dataset, per-classifier time-series (ICU) downstream utility, AUROC (with the clinical-validity layer). \utilitycolumnkey}\label{tab:supp-util-ts-auroc}\\
\toprule
\textbf{Dataset} & \textbf{Task} & \textbf{Classifier} & \textbf{Paper} & \textbf{Real} & \textbf{CTGAN} & \textbf{TVAE} & \textbf{CG} & \textbf{GC} \\
\midrule\endfirsthead
\multicolumn{9}{@{}l}{\emph{\tablename~\thetable\ (continued)}}\\
\toprule
\textbf{Dataset} & \textbf{Task} & \textbf{Classifier} & \textbf{Paper} & \textbf{Real} & \textbf{CTGAN} & \textbf{TVAE} & \textbf{CG} & \textbf{GC} \\
\midrule\endhead
\midrule\multicolumn{9}{r@{}}{\emph{continued on next page}}\\\endfoot
\bottomrule\endlastfoot
MIMIC-III & HFM & LR & 0.842 & 0.777 & 0.533 (68.5\%) & 0.816 (105.0\%) & 0.556 (71.5\%) & 0.824 (106.0\%) \\*
MIMIC-III & HFM & RF & -- & 0.858 & 0.562 (65.5\%) & 0.852 (99.3\%) & 0.462 (53.9\%) & 0.763 (89.0\%) \\*
MIMIC-III & HFM & GB & -- & 0.802 & 0.339 (42.3\%) & 0.785 (97.9\%) & 0.406 (50.6\%) & 0.717 (89.4\%) \\*
MIMIC-III & HFM & XGB & -- & 0.815 & 0.578 (71.0\%) & 0.810 (99.5\%) & 0.466 (57.3\%) & 0.719 (88.3\%) \\*
MIMIC-III & HFM & MLP & -- & 0.472 & 0.388 (82.2\%) & 0.787 (166.7\%) & 0.417 (88.4\%) & 0.753 (159.4\%) \\
MIMIC-III & Hosp-R & LR & 0.580 & 0.611 & 0.533 (87.3\%) & 0.608 (99.6\%) & 0.544 (89.1\%) & 0.600 (98.3\%) \\*
MIMIC-III & Hosp-R & RF & 0.680 & 0.616 & 0.522 (84.9\%) & 0.622 (101.0\%) & 0.499 (81.1\%) & 0.484 (78.6\%) \\*
MIMIC-III & Hosp-R & GB & 0.720 & 0.606 & 0.545 (90.0\%) & 0.602 (99.4\%) & 0.533 (88.1\%) & 0.531 (87.7\%) \\
\end{longtable}
\endgroup
\par\medskip

\begingroup
\scriptsize\setlength{\tabcolsep}{2pt}\renewcommand{\arraystretch}{1.0}
\Needspace{11\baselineskip}
\begin{longtable}{@{}l l l c c c c c c@{}}
\caption{Per-dataset, per-classifier time-series (ICU) downstream utility, AUPRC (with the clinical-validity layer). \utilitycolumnkey}\label{tab:supp-util-ts-auprc}\\
\toprule
\textbf{Dataset} & \textbf{Task} & \textbf{Classifier} & \textbf{Paper} & \textbf{Real} & \textbf{CTGAN} & \textbf{TVAE} & \textbf{CG} & \textbf{GC} \\
\midrule\endfirsthead
\multicolumn{9}{@{}l}{\emph{\tablename~\thetable\ (continued)}}\\
\toprule
\textbf{Dataset} & \textbf{Task} & \textbf{Classifier} & \textbf{Paper} & \textbf{Real} & \textbf{CTGAN} & \textbf{TVAE} & \textbf{CG} & \textbf{GC} \\
\midrule\endhead
\midrule\multicolumn{9}{r@{}}{\emph{continued on next page}}\\\endfoot
\bottomrule\endlastfoot
MIMIC-III & HFM & LR & -- & 0.413 & 0.136 (32.9\%) & 0.459 (111.1\%) & 0.180 (43.5\%) & 0.411 (99.4\%) \\*
MIMIC-III & HFM & RF & -- & 0.548 & 0.136 (24.9\%) & 0.493 (90.0\%) & 0.110 (20.1\%) & 0.336 (61.4\%) \\*
MIMIC-III & HFM & GB & -- & 0.509 & 0.089 (17.4\%) & 0.415 (81.6\%) & 0.102 (20.1\%) & 0.316 (62.1\%) \\*
MIMIC-III & HFM & XGB & -- & 0.475 & 0.149 (31.4\%) & 0.465 (97.9\%) & 0.114 (24.0\%) & 0.290 (61.1\%) \\*
MIMIC-III & HFM & MLP & -- & 0.122 & 0.123 (100.6\%) & 0.435 (355.8\%) & 0.110 (89.7\%) & 0.325 (265.5\%) \\
MIMIC-III & Hosp-R & LR & -- & 0.078 & 0.063 (80.6\%) & 0.077 (98.9\%) & 0.064 (82.8\%) & 0.070 (90.6\%) \\*
MIMIC-III & Hosp-R & RF & -- & 0.079 & 0.056 (70.9\%) & 0.082 (104.2\%) & 0.051 (64.2\%) & 0.054 (68.6\%) \\*
MIMIC-III & Hosp-R & GB & -- & 0.077 & 0.065 (84.5\%) & 0.079 (103.0\%) & 0.060 (77.5\%) & 0.058 (75.4\%) \\
\end{longtable}
\endgroup
\par\medskip

\Needspace{12\baselineskip}
\section{Multi-class Classification Results}
\noindent Three tasks in the benchmark are multi-class: SUPPORT2 FO (5 classes), UCI-Derm (6 classes), and UCI Diabetes~130 Readmission (3 classes). Abbreviations are defined in Table~\ref{tab:abbr-key}. For these tasks AUROC is the macro one-vs-rest AUROC and AUPRC is the macro AUPRC, computed across all classes.\par\medskip
\begingroup
\scriptsize\setlength{\tabcolsep}{2pt}\renewcommand{\arraystretch}{1.0}
\Needspace{11\baselineskip}
\begin{longtable}{@{}l l l c c c c c c@{}}
\caption{Multi-class tabular downstream utility, macro one-vs-rest AUROC (per classifier). \utilitycolumnkey}\label{tab:supp-mc-auroc}\\
\toprule
\textbf{Dataset} & \textbf{Task} & \textbf{Classifier} & \textbf{Paper} & \textbf{Real} & \textbf{CTGAN} & \textbf{TVAE} & \textbf{CG} & \textbf{GC} \\
\midrule\endfirsthead
\multicolumn{9}{@{}l}{\emph{\tablename~\thetable\ (continued)}}\\
\toprule
\textbf{Dataset} & \textbf{Task} & \textbf{Classifier} & \textbf{Paper} & \textbf{Real} & \textbf{CTGAN} & \textbf{TVAE} & \textbf{CG} & \textbf{GC} \\
\midrule\endhead
\midrule\multicolumn{9}{r@{}}{\emph{continued on next page}}\\\endfoot
\bottomrule\endlastfoot
SUPPORT2 & FO & LR & -- & 0.807 & 0.555 (68.7\%) & 0.748 (92.7\%) & 0.555 (68.7\%) & 0.567 (70.3\%) \\*
SUPPORT2 & FO & RF & -- & 0.821 & 0.607 (73.9\%) & 0.773 (94.2\%) & 0.598 (72.8\%) & 0.612 (74.5\%) \\*
SUPPORT2 & FO & GB & -- & 0.833 & 0.602 (72.3\%) & 0.780 (93.5\%) & 0.600 (72.0\%) & 0.603 (72.4\%) \\*
SUPPORT2 & FO & XGB & -- & 0.823 & 0.580 (70.5\%) & 0.774 (94.0\%) & 0.574 (69.7\%) & 0.589 (71.6\%) \\*
SUPPORT2 & FO & MLP & -- & 0.826 & 0.545 (66.0\%) & 0.782 (94.6\%) & 0.554 (67.0\%) & 0.568 (68.8\%) \\
UCI-Derm & DDC & LR & -- & 0.999 & 0.834 (83.4\%) & 0.995 (99.5\%) & 0.853 (85.3\%) & 0.781 (78.2\%) \\*
UCI-Derm & DDC & RF & -- & 0.998 & 0.872 (87.4\%) & 1.000 (100.1\%) & 0.856 (85.7\%) & 0.779 (78.1\%) \\*
UCI-Derm & DDC & GB & -- & 0.995 & 0.872 (87.7\%) & 0.997 (100.2\%) & 0.871 (87.6\%) & 0.709 (71.2\%) \\
\end{longtable}
\endgroup
\par\medskip

\begingroup
\scriptsize\setlength{\tabcolsep}{2pt}\renewcommand{\arraystretch}{1.0}
\Needspace{11\baselineskip}
\begin{longtable}{@{}l l l c c c c c c@{}}
\caption{Multi-class tabular downstream utility, macro AUPRC (per classifier). \utilitycolumnkey}\label{tab:supp-mc-auprc}\\
\toprule
\textbf{Dataset} & \textbf{Task} & \textbf{Classifier} & \textbf{Paper} & \textbf{Real} & \textbf{CTGAN} & \textbf{TVAE} & \textbf{CG} & \textbf{GC} \\
\midrule\endfirsthead
\multicolumn{9}{@{}l}{\emph{\tablename~\thetable\ (continued)}}\\
\toprule
\textbf{Dataset} & \textbf{Task} & \textbf{Classifier} & \textbf{Paper} & \textbf{Real} & \textbf{CTGAN} & \textbf{TVAE} & \textbf{CG} & \textbf{GC} \\
\midrule\endhead
\midrule\multicolumn{9}{r@{}}{\emph{continued on next page}}\\\endfoot
\bottomrule\endlastfoot
SUPPORT2 & FO & LR & -- & 0.449 & 0.287 (64.0\%) & 0.400 (89.2\%) & 0.298 (66.3\%) & 0.301 (67.0\%) \\*
SUPPORT2 & FO & RF & -- & 0.454 & 0.275 (60.7\%) & 0.413 (91.2\%) & 0.310 (68.4\%) & 0.321 (70.7\%) \\*
SUPPORT2 & FO & GB & -- & 0.469 & 0.257 (54.7\%) & 0.410 (87.5\%) & 0.301 (64.1\%) & 0.312 (66.5\%) \\*
SUPPORT2 & FO & XGB & -- & 0.459 & 0.256 (55.8\%) & 0.399 (86.9\%) & 0.296 (64.6\%) & 0.303 (66.0\%) \\*
SUPPORT2 & FO & MLP & -- & 0.462 & 0.296 (64.1\%) & 0.410 (88.6\%) & 0.311 (67.3\%) & 0.315 (68.2\%) \\
UCI-Derm & DDC & LR & -- & 0.997 & 0.629 (63.1\%) & 0.978 (98.2\%) & 0.508 (51.0\%) & 0.513 (51.5\%) \\*
UCI-Derm & DDC & RF & -- & 0.991 & 0.604 (61.0\%) & 0.998 (100.7\%) & 0.591 (59.6\%) & 0.472 (47.7\%) \\*
UCI-Derm & DDC & GB & -- & 0.976 & 0.607 (62.2\%) & 0.984 (100.8\%) & 0.596 (61.1\%) & 0.447 (45.8\%) \\
\end{longtable}
\endgroup
\par\medskip

\Needspace{12\baselineskip}
\section{Ablation: Effect of the Clinical-Validity Layer}
\noindent To isolate the contribution of the model-agnostic clinical-validity layer, we re-run the utility evaluation with the layer disabled and report AUROC retention. Only datasets that carry a validity requirement are included; comparing against the main utility tables quantifies how much downstream signal is attributable to enforcing clinical and structural validity.\par\medskip
\begingroup
\scriptsize\setlength{\tabcolsep}{2pt}\renewcommand{\arraystretch}{1.0}
\Needspace{11\baselineskip}
\begin{longtable}{@{}l l l c c c c c c@{}}
\caption{Tabular ablation: downstream utility (AUROC, Retention) with the clinical-validity layer \emph{disabled}. \utilitycolumnkey}\label{tab:supp-abl-tab}\\
\toprule
\textbf{Dataset} & \textbf{Task} & \textbf{Classifier} & \textbf{Paper} & \textbf{Real} & \textbf{CTGAN} & \textbf{TVAE} & \textbf{CG} & \textbf{GC} \\
\midrule\endfirsthead
\multicolumn{9}{@{}l}{\emph{\tablename~\thetable\ (continued)}}\\
\toprule
\textbf{Dataset} & \textbf{Task} & \textbf{Classifier} & \textbf{Paper} & \textbf{Real} & \textbf{CTGAN} & \textbf{TVAE} & \textbf{CG} & \textbf{GC} \\
\midrule\endhead
\midrule\multicolumn{9}{r@{}}{\emph{continued on next page}}\\\endfoot
\bottomrule\endlastfoot
SUPPORT2 & Death & LR & -- & 0.841 & 0.788 (93.8\%) & 0.820 (97.6\%) & 0.809 (96.2\%) & 0.792 (94.2\%) \\*
SUPPORT2 & Death & RF & -- & 0.849 & 0.783 (92.2\%) & 0.825 (97.2\%) & 0.814 (95.9\%) & 0.786 (92.6\%) \\*
SUPPORT2 & Death & GB & -- & 0.855 & 0.789 (92.2\%) & 0.827 (96.6\%) & 0.809 (94.6\%) & 0.776 (90.7\%) \\*
SUPPORT2 & Death & XGB & -- & 0.847 & 0.763 (90.0\%) & 0.828 (97.8\%) & 0.791 (93.4\%) & 0.763 (90.1\%) \\*
SUPPORT2 & Death & MLP & -- & 0.846 & 0.726 (85.8\%) & 0.824 (97.3\%) & 0.801 (94.7\%) & 0.759 (89.7\%) \\
SUPPORT2 & FO & LR & -- & 0.807 & 0.762 (94.3\%) & 0.817 (101.2\%) & 0.706 (87.4\%) & 0.576 (71.3\%) \\*
SUPPORT2 & FO & RF & -- & 0.821 & 0.757 (92.2\%) & 0.819 (99.8\%) & 0.767 (93.5\%) & 0.570 (69.5\%) \\*
SUPPORT2 & FO & GB & -- & 0.833 & 0.763 (91.5\%) & 0.824 (98.9\%) & 0.766 (91.9\%) & 0.599 (71.9\%) \\*
SUPPORT2 & FO & XGB & -- & 0.823 & 0.762 (92.6\%) & 0.838 (101.7\%) & 0.742 (90.2\%) & 0.593 (72.1\%) \\*
SUPPORT2 & FO & MLP & -- & 0.826 & 0.669 (81.0\%) & 0.826 (100.0\%) & 0.714 (86.5\%) & 0.557 (67.4\%) \\
SUPPORT2 & Mortality & LR & -- & 0.946 & 0.929 (98.2\%) & 0.943 (99.7\%) & 0.929 (98.2\%) & 0.919 (97.1\%) \\*
SUPPORT2 & Mortality & RF & -- & 0.949 & 0.933 (98.3\%) & 0.942 (99.2\%) & 0.929 (98.0\%) & 0.879 (92.6\%) \\*
SUPPORT2 & Mortality & GB & -- & 0.951 & 0.925 (97.3\%) & 0.944 (99.2\%) & 0.927 (97.5\%) & 0.854 (89.8\%) \\*
SUPPORT2 & Mortality & XGB & -- & 0.950 & 0.910 (95.8\%) & 0.946 (99.6\%) & 0.918 (96.6\%) & 0.748 (78.7\%) \\*
SUPPORT2 & Mortality & MLP & -- & 0.951 & 0.927 (97.6\%) & 0.942 (99.1\%) & 0.927 (97.5\%) & 0.883 (92.8\%) \\
UCI-CC & CC & LR & -- & 0.574 & 0.392 (68.3\%) & 0.473 (82.4\%) & 0.452 (78.7\%) & 0.385 (67.0\%) \\*
UCI-CC & CC & RF & -- & 0.698 & 0.577 (82.7\%) & 0.580 (83.1\%) & 0.411 (58.9\%) & 0.643 (92.1\%) \\*
UCI-CC & CC & GB & -- & 0.665 & 0.430 (64.6\%) & 0.600 (90.2\%) & 0.403 (60.5\%) & 0.468 (70.3\%) \\*
UCI-CC & CC & XGB & -- & 0.651 & 0.531 (81.6\%) & 0.504 (77.5\%) & 0.289 (44.4\%) & 0.432 (66.4\%) \\*
UCI-CC & CC & MLP & -- & 0.427 & 0.452 (105.8\%) & 0.526 (123.3\%) & 0.461 (108.1\%) & 0.374 (87.7\%) \\
\end{longtable}
\endgroup
\par\medskip

\begingroup
\scriptsize\setlength{\tabcolsep}{2pt}\renewcommand{\arraystretch}{1.0}
\Needspace{11\baselineskip}
\begin{longtable}{@{}l l l c c c c c c@{}}
\caption{Time-series ablation: downstream utility (AUROC, Retention) with the clinical-validity layer \emph{disabled}. \utilitycolumnkey}\label{tab:supp-abl-ts}\\
\toprule
\textbf{Dataset} & \textbf{Task} & \textbf{Classifier} & \textbf{Paper} & \textbf{Real} & \textbf{CTGAN} & \textbf{TVAE} & \textbf{CG} & \textbf{GC} \\
\midrule\endfirsthead
\multicolumn{9}{@{}l}{\emph{\tablename~\thetable\ (continued)}}\\
\toprule
\textbf{Dataset} & \textbf{Task} & \textbf{Classifier} & \textbf{Paper} & \textbf{Real} & \textbf{CTGAN} & \textbf{TVAE} & \textbf{CG} & \textbf{GC} \\
\midrule\endhead
\midrule\multicolumn{9}{r@{}}{\emph{continued on next page}}\\\endfoot
\bottomrule\endlastfoot
MIMIC-III & HFM & LR & 0.842 & 0.777 & 0.637 (82.0\%) & 0.863 (111.0\%) & 0.622 (80.0\%) & 0.824 (106.0\%) \\*
MIMIC-III & HFM & RF & -- & 0.858 & 0.488 (56.9\%) & 0.864 (100.8\%) & 0.461 (53.8\%) & 0.763 (89.0\%) \\*
MIMIC-III & HFM & GB & -- & 0.802 & 0.365 (45.5\%) & 0.840 (104.7\%) & 0.517 (64.4\%) & 0.717 (89.4\%) \\*
MIMIC-III & HFM & XGB & -- & 0.815 & 0.465 (57.0\%) & 0.868 (106.5\%) & 0.560 (68.8\%) & 0.719 (88.3\%) \\*
MIMIC-III & HFM & MLP & -- & 0.472 & 0.536 (113.4\%) & 0.829 (175.6\%) & 0.452 (95.7\%) & 0.753 (159.4\%) \\
MIMIC-III & LOS & LR & -- & 0.739 & 0.675 (91.3\%) & 0.710 (96.1\%) & 0.682 (92.3\%) & 0.696 (94.3\%) \\*
MIMIC-III & LOS & RF & -- & 0.765 & 0.691 (90.3\%) & 0.724 (94.6\%) & 0.696 (91.0\%) & 0.565 (73.9\%) \\*
MIMIC-III & LOS & GB & -- & 0.772 & 0.701 (90.8\%) & 0.732 (94.7\%) & 0.698 (90.3\%) & 0.650 (84.2\%) \\*
MIMIC-III & LOS & XGB & -- & 0.764 & 0.636 (83.2\%) & 0.746 (97.6\%) & 0.623 (81.5\%) & 0.517 (67.7\%) \\*
MIMIC-III & LOS & MLP & -- & 0.746 & 0.648 (86.8\%) & 0.725 (97.2\%) & 0.644 (86.3\%) & 0.553 (74.0\%) \\
MIMIC-IV & HFM & LR & -- & 0.821 & 0.766 (93.2\%) & 0.788 (96.0\%) & 0.770 (93.8\%) & 0.715 (87.0\%) \\*
MIMIC-IV & HFM & RF & -- & 0.829 & 0.758 (91.4\%) & 0.781 (94.2\%) & 0.766 (92.4\%) & 0.676 (81.5\%) \\*
MIMIC-IV & HFM & GB & -- & 0.839 & 0.767 (91.3\%) & 0.792 (94.4\%) & 0.775 (92.3\%) & 0.625 (74.5\%) \\*
MIMIC-IV & HFM & XGB & 0.820 & 0.828 & 0.709 (85.7\%) & 0.808 (97.6\%) & 0.750 (90.6\%) & 0.566 (68.3\%) \\*
MIMIC-IV & HFM & MLP & -- & 0.815 & 0.691 (84.8\%) & 0.787 (96.5\%) & 0.732 (89.8\%) & 0.467 (57.3\%) \\
MIMIC-IV & LOS & LR & -- & 0.815 & 0.799 (98.1\%) & 0.741 (91.0\%) & 0.799 (98.1\%) & 0.779 (95.7\%) \\*
MIMIC-IV & LOS & RF & -- & 0.831 & 0.789 (95.0\%) & 0.752 (90.5\%) & 0.789 (95.0\%) & 0.740 (89.1\%) \\*
MIMIC-IV & LOS & GB & -- & 0.829 & 0.806 (97.2\%) & 0.763 (92.0\%) & 0.804 (97.0\%) & 0.766 (92.3\%) \\*
MIMIC-IV & LOS & XGB & 0.860 & 0.835 & 0.773 (92.6\%) & 0.744 (89.1\%) & 0.776 (92.9\%) & 0.712 (85.3\%) \\*
MIMIC-IV & LOS & MLP & -- & 0.831 & 0.792 (95.2\%) & 0.727 (87.4\%) & 0.790 (95.1\%) & 0.752 (90.5\%) \\
eICU & HFM & LR & -- & 0.747 & 0.718 (96.1\%) & 0.748 (100.1\%) & 0.723 (96.7\%) & 0.726 (97.2\%) \\*
eICU & HFM & RF & -- & 0.753 & 0.668 (88.7\%) & 0.731 (97.0\%) & 0.694 (92.1\%) & 0.691 (91.7\%) \\*
eICU & HFM & GB & -- & 0.750 & 0.663 (88.4\%) & 0.746 (99.5\%) & 0.705 (94.1\%) & 0.676 (90.2\%) \\*
eICU & HFM & XGB & -- & 0.719 & 0.613 (85.3\%) & 0.753 (104.8\%) & 0.642 (89.4\%) & 0.611 (85.0\%) \\*
eICU & HFM & MLP & -- & 0.724 & 0.677 (93.5\%) & 0.732 (101.1\%) & 0.656 (90.7\%) & 0.616 (85.1\%) \\
eICU & LOS & LR & -- & 0.746 & 0.738 (98.9\%) & 0.703 (94.2\%) & 0.733 (98.2\%) & 0.711 (95.3\%) \\*
eICU & LOS & RF & -- & 0.762 & 0.724 (95.0\%) & 0.690 (90.6\%) & 0.730 (95.8\%) & 0.611 (80.1\%) \\*
eICU & LOS & GB & -- & 0.763 & 0.746 (97.8\%) & 0.718 (94.1\%) & 0.743 (97.5\%) & 0.690 (90.5\%) \\*
eICU & LOS & XGB & -- & 0.747 & 0.695 (93.0\%) & 0.709 (94.9\%) & 0.699 (93.6\%) & 0.566 (75.7\%) \\*
eICU & LOS & MLP & -- & 0.757 & 0.734 (96.9\%) & 0.650 (85.8\%) & 0.730 (96.4\%) & 0.609 (80.4\%) \\
\end{longtable}
\endgroup
\par\medskip

\section{Implementation Details: Hyperparameters}

\noindent All synthetic data generation and downstream evaluation were run on Google Colab with a single NVIDIA L4 GPU.   

\subsection{Synthetic Data Generators}
\noindent All four CoMed generators are built on the corresponding SDV single-table synthesizers. The following settings are shared across every dataset and generator: random \texttt{seed}$=42$; GPU used when available (\texttt{cuda}); domain-validity constraints applied during fitting; \texttt{enforce\_min\_max\_values}$=$True and \texttt{enforce\_rounding}$=$False; categorical columns with at least $20$ distinct values re-encoded with an \texttt{rdt} \texttt{UniformEncoder} before fitting; and one synthetic dataset generated at the same row count as the real training split.

\begin{itemize}
\item \textbf{CoMed-CTGAN} (\texttt{CTGANSynthesizer}): \texttt{generator\_lr}$=$\texttt{discriminator\_lr}$=2\times10^{-4}$; \texttt{generator\_decay}$=$\texttt{discriminator\_decay}$=10^{-6}$; \texttt{discriminator\_steps}$=1$; \texttt{pac}$=10$. Tabular: \texttt{embedding\_dim}$=128$, \texttt{generator\_dim}$=$\texttt{discriminator\_dim}$=(256,256)$. Time-series: \texttt{embedding\_dim}$=512$, \texttt{generator\_dim}$=$\texttt{discriminator\_dim}$=(256,512,256)$.
\item \textbf{CoMed-TVAE} (\texttt{TVAESynthesizer}): \texttt{embedding\_dim}, \texttt{compress\_dims} and \texttt{decompress\_dims} mirror the CTGAN dimensions (tabular $128/(256,256)$; time-series $512/(256,512,256)$).
\item \textbf{CoMed-CopulaGAN} (\texttt{CopulaGANSynthesizer}): \texttt{embedding\_dim}, \texttt{generator\_dim}, \texttt{discriminator\_dim}, and \texttt{pac}$=10$ as in CTGAN.
\item \textbf{CoMed-GaussianCopula} (\texttt{GaussianCopulaSynthesizer}): statistical model with no epochs, batch size, or network dimensions; uses only the shared constraint and encoding settings above.
\end{itemize}

\noindent Training epochs and batch size scale with dataset size and are identical across CTGAN, TVAE, and CopulaGAN for a given dataset (GaussianCopula ignores them). For \emph{tabular} data, \texttt{total\_epochs} ranges from $300$ (largest datasets, $\geq 100$k rows) through $500$--$1200$ (mid-size) to $1500$ (small datasets), and \texttt{batch\_size} ranges from $50$ to $2000$ (chosen even, divisible by \texttt{pac}$=10$, and no larger than the row count). For \emph{time-series} data, \texttt{total\_epochs} is $1500$ ($2000$ for the readmission tasks) and \texttt{batch\_size} ranges from $500$ to $5000$, scaled to the number of rows.

\subsection{Downstream Classifiers (Tabular)}
\begin{itemize}
\item \textbf{Logistic Regression}: \texttt{max\_iter}$=2000$, \texttt{class\_weight}$=$``balanced'', \texttt{solver}$=$``lbfgs'', \texttt{random\_state}$=42$.
\item \textbf{Random Forest}: \texttt{n\_estimators}$=300$, \texttt{class\_weight}$=$``balanced'', \texttt{n\_jobs}$=-1$, \texttt{random\_state}$=42$.
\item \textbf{Gradient Boosting}: \texttt{n\_estimators}$=100$ ($60$ for very large datasets), \texttt{subsample}$=1.0$ ($0.6$ for very large datasets), \texttt{max\_depth}$=3$, \texttt{random\_state}$=42$.
\item \textbf{XGBoost}: \texttt{n\_estimators}$=300$, \texttt{max\_depth}$=6$, \texttt{learning\_rate}$=0.1$, \texttt{subsample}$=0.9$, \texttt{colsample\_bytree}$=0.9$, \texttt{eval\_metric}$=$``logloss'', \texttt{tree\_method}$=$``hist'', \texttt{n\_jobs}$=-1$, \texttt{random\_state}$=42$.
\item \textbf{MLP}: \texttt{hidden\_layer\_sizes}$=(100,)$ (or $(128,64)$ for larger datasets), \texttt{max\_iter}$=300$ ($120$ for large datasets), \texttt{early\_stopping}$=$True, \texttt{n\_iter\_no\_change}$=15$, \texttt{random\_state}$=42$.
\end{itemize}
\noindent Evaluation uses a stratified $80/20$ train/test split with \texttt{random\_state}$=42$.

\subsection{Downstream Classifiers (Time-Series)}
\begin{itemize}
\item \textbf{Logistic Regression}: \texttt{max\_iter}$=2000$, \texttt{class\_weight}$=$``balanced'', \texttt{solver}$=$``lbfgs'', \texttt{random\_state}$=42$.
\item \textbf{Random Forest}: \texttt{n\_estimators}$=400$, \texttt{class\_weight}$=$``balanced'', \texttt{n\_jobs}$=-1$, \texttt{random\_state}$=42$.
\item \textbf{Gradient Boosting}: \texttt{random\_state}$=42$ (defaults otherwise).
\item \textbf{XGBoost}: \texttt{n\_estimators}$=400$, \texttt{max\_depth}$=6$, \texttt{learning\_rate}$=0.1$, \texttt{subsample}$=0.9$, \texttt{colsample\_bytree}$=0.9$, \texttt{eval\_metric}$=$``logloss'', \texttt{scale\_pos\_weight} computed from class counts, \texttt{tree\_method}$=$``hist'', \texttt{n\_jobs}$=-1$, \texttt{random\_state}$=42$.
\item \textbf{MLP}: \texttt{hidden\_layer\_sizes}$=(128,64)$, \texttt{max\_iter}$=500$, \texttt{early\_stopping}$=$True, \texttt{n\_iter\_no\_change}$=15$, \texttt{validation\_fraction}$=0.1$, \texttt{random\_state}$=42$.
\end{itemize}
\noindent The split and evaluation setup use \texttt{random\_seed}$=42$.